\documentclass{article} 
\usepackage{iclr2023_conference, times}
\pdfoutput=1

\usepackage{amsmath,amsfonts,bm}

\def\eqref#1{equation~\ref{#1}}

\def\1{\bm{1}}

\DeclareMathAlphabet{\mathsfit}{\encodingdefault}{\sfdefault}{m}{sl}
\SetMathAlphabet{\mathsfit}{bold}{\encodingdefault}{\sfdefault}{bx}{n}

\usepackage{graphicx}
\usepackage{hyperref}
\usepackage{url}
\usepackage{tabularx}
\usepackage{booktabs}
\usepackage{wrapfig}
\usepackage{comment} 

\usepackage{multirow}

\title{ESB: A Benchmark For Multi-Domain \\ End-to-End Speech Recognition}

\author{Sanchit Gandhi, Patrick von Platen \& Alexander M. Rush \\
Hugging Face \\
\texttt{\{sanchit, patrick, sasha\}@huggingface.co} \\
}

\iclrfinalcopy 

\renewcommand{\headrulewidth}{0pt}
\renewcommand{\footrulewidth}{0pt}

\begin{document}

\maketitle

\begin{abstract}

Speech recognition applications cover a range of different audio and text distributions, with different speaking styles, background noise, transcription punctuation and character casing. However, many speech recognition systems require dataset-specific tuning (audio filtering, punctuation removal and normalisation of casing), therefore assuming a-priori knowledge of both the audio and text distributions. This tuning requirement can lead to systems failing to generalise to other datasets and domains. To promote the development of multi-domain speech systems, we introduce the End-to-end Speech Benchmark (ESB) for evaluating the performance of a single automatic speech recognition (ASR) system across a broad set of speech datasets. Benchmarked systems must use the same data pre- and post-processing algorithm across datasets - assuming the audio and text data distributions are a-priori unknown. We compare a series of state-of-the-art (SoTA) end-to-end (E2E) systems on this benchmark, demonstrating how a single speech system can be applied and evaluated on a wide range of data distributions. We find E2E systems to be effective across datasets: in a fair comparison, E2E systems achieve within 2.6\% of SoTA systems tuned to a specific dataset. Our analysis reveals that transcription artefacts, such as punctuation and casing, pose difficulties for ASR systems and should be included in evaluation. We believe E2E benchmarking over a range of datasets promotes the research of multi-domain speech recognition systems. ESB is available at \url{https://huggingface.co/esb}.

\end{abstract}

\section{Introduction}
Speech recognition covers various applications, including dictation, voice assistants, video captioning, telephone conversations and meeting transcriptions \citep{aksenova21_asrbenchmarks}. Each application has domain-specific data distributions for both the audio inputs and transcription outputs. The audio inputs are derived from different recording conditions, degrees of background noise, speakers and styles (narrated, oratory or spontaneous).
The nature of the transcriptions is also domain-dependent; in formal settings, such as meeting transcriptions, the text must be orthographic\footnote{\emph{orthographic}: the accepted way of spelling and writing words according to standard usage \citep{mcintosh15_dictionary}.} and satisfy standard formatting conventions. Whereas in more informal settings, such as telephone conversations, punctuation and casing are often omitted \citep{KIM2003563}. To handle the diversity of speech recognition conditions, there is a need for multi-domain systems that maintain their performance over a collection of datasets with different audio and transcription distributions.

However, most automatic speech recognition (ASR) systems are trained and evaluated on a single dataset, utilising dataset-specific model architectures and pre-/post-processing to optimise for single dataset performance \citep{likhomanenko20_arxiv}. Such dataset-specific tuning assumes a-priori knowledge of both the audio and text distribution and yields systems that transfer poorly to other datasets and domains. A generalisable system should transfer to different datasets and domains with training data, but without the need for dataset-specific tuning \citep{wang19_glue} or a-priori knowledge of the data distributions. End-to-end (E2E) systems consist of a single model that maps the raw audio inputs to the transcription outputs \citep{graves14_towards}. Learning directly from data, E2E systems do not require dataset-specific configurations \citep{hannun14_deepspeech}. As such, they can be applied independently to different datasets and domains \citep{chan21_speechstew, radford22_whipser}.

To facilitate the research of multi-domain, generalisable ASR systems, we present the End-to-end Speech Benchmark (ESB), a benchmark for evaluating a single ASR system across a collection of speech datasets spanning different domains and speech recognition conditions. Benchmarked systems must use the same architecture across datasets and may not use dataset-specific pre- or post-processing. Therefore, ESB favours systems that can be applied independently across speech recognition domains with no a-priori knowledge of the data distributions. None of the datasets presented in ESB were created specifically for the benchmark; all datasets are pre-existing for the reason that they are widely considered by the speech community to be the most applicable, challenging and interesting datasets. We adopt an open-source and open-science approach by considering datasets that are freely available and accessible.


To demonstrate ESB, we perform baseline experiments with five different E2E approaches. We find these E2E systems to be effective across datasets. In a fair comparison, they perform to within 2.6\% word error rate of state-of-the-art systems tuned to a specific dataset. Our analysis shows that transcription artefacts, such as punctuation and casing, make the task of speech recognition more difficult and should be included in evaluation. We believe E2E benchmarking over a range of datasets encourages the research of multi-domain speech recognition systems.

\section{Related Work}

Speech recognition datasets have long focused on covering different domains and speaking styles: the TIMIT \citep{garofolo93_darpa} and Wall-Street Journal \citep{ldc93_wsj} corpora contain news broadcast recordings, SwitchBoard \citep{godfrey92_switchboard} and Fisher \citep{ldc04_fisher_speech, ldc04_fisher_transcriptions, ldc04_fisher_speech_2, ldc04_fisher_transcriptions_2} spontaneous telephone conversations, LibriSpeech \citep{panayotov15_libripseech} narrated audiobooks, Common Voice \citep{ardila20_commonvoice} narrated Wikipedia articles and TED-LIUM \citep{hernandez18_tedlium} oratory educational talks. More recently, datasets such as People's Speech \citep{galvez21_peoples_speech} and GigaSpeech \citep{chen21_gigaspeech} extend this to cover multiple domains in one dataset. However, these datasets lack certain important domains and speaking styles, such as conversational speech, which are currently only covered by certain individual datasets. 
We see this as an important trend towards multi-domain speech recognition and collect different datasets to form a unified ASR benchmark.

Traditionally, ASR systems are trained on case and punctuation normalised text \citep{nist98_snor, Povey11_kaldi}; the transcriptions are pre-processed to remove casing and punctuation before training and evaluation. However, in certain speech recognition applications, orthographic transcriptions are required \cite{kim01_prosody}. 
Recent work has looked at training ASR systems on orthographic transcriptions \citep{oneill21_kensho, radford22_whipser}, relying on a data-driven E2E approach in learning to predict cased and punctuated outputs. However, the features of orthographic text remain challenging for ASR systems. We evaluate a single system over multiple datasets and include all dataset-specific transcription formatting requirements.


For text understanding, GLUE \citep{wang19_glue} and SuperGLUE \citep{wang19_superglue} provide well established benchmarks for assessing the generalisation abilities of a single system over a range of different natural language understanding tasks. The SUPERB \citep{yang21_superb} and XTREME-S \citep{conneau22_xtreme_s} benchmarks assess a single system over a multiple spoken language processing tasks. This paper extends these efforts to show that English ASR has sufficient diversity in datasets and domains to merit a benchmark of its own.

\section{Motivation for an End-to-End Benchmark}
Different speech domains have different data distributions for audio artefacts (quality, speakers and styles) and transcription outputs (punctuation, casing, orthography). In using the term \textit{end-to-end} (E2E), we refer to systems that map from the raw audio inputs to the transcription outputs without domain-specific architectures or additional processing. In this section, we describe the existing works regarding multi-domain and E2E ASR and outline the principal issues involved.

Recent datasets have focused on domains with more challenging audio inputs, specifically in audio quality, speakers and speaking style \citep{panayotov15_libripseech, ardila20_commonvoice, wang21_voxpopuli, hernandez18_tedlium, chen21_gigaspeech, oneill21_kensho, delrio22_earnings22, carletta07_ami, renals07_ami, godfrey92_switchboard, ldc04_fisher_speech, ldc04_fisher_transcriptions, ldc04_fisher_speech_2, ldc04_fisher_transcriptions_2}. These datasets incorporate distinct audio domains, each with different recording conditions and degrees of background noise. Each dataset includes speakers from both native or non-native English speaking backgrounds, and together cover accents and dialects from seven different language regions \citep{delrio22_earnings22}. The speaking style for each dataset falls into one of three categories: narrated, oratory or spontaneous, with each style having different distributions for speaking speed and utterance length. We discuss the individual datasets in detail in Section \ref{sec:datasets}.



For many ASR systems, a series of dataset specific pre- and post-processing steps are applied when training and evaluating systems on individual datasets. For the 10 datasets in this work, there are 10 different Kaldi \citep{Povey11_kaldi} recipes in use, each with unique pre- and post-processing steps. Of these recipes, one is not even publicly accessible.
Employing dataset-specific pre-processing steps results in systems that do not transfer to different domains. For example, a system that extracts speech features without a noise-suppression algorithm works adequately well for a dataset with low-background noise, but the same approach produces much worse results on a noisy dataset \citep{kim16_pncc}.

Recent speech recognition datasets also include full transcriptions with all the necessary orthographic features required for their respective domains \citep{carletta07_ami, renals07_ami, oneill21_kensho, delrio22_earnings22}. These datasets aim to encourage ASR systems capable of producing transcriptions that adhere to the formatting requirements of the target text domain.
We note that this differs from the standard ASR output transcription format known as Standard Normalised Orthographic Representation (SNOR) \citep{nist98_snor}, which consists of single-case letters without punctuation marks or numbers. This format is necessary for ASR systems that do not predict punctuated and cased outputs, relying on post-processing to restore transcription formatting \citep{chen1999_speech}. Per contra, many speech recognition applications, such as financial meeting transcriptions or legal documents, require orthographic text. 


In circumstances where orthographic text is required, it is typically achieved through a series of dataset-specific post-processing steps applied to the ASR output, each of which treats a single orthographic feature \citep{beeferman98_cyberpunc, lita_03_truecaseing, KIM2003563, gravano09_restoring, yuan-briscoe-2016-grammatical}. However, there are significant shortcomings to this pipeline approach. Firstly, certain orthographic decisions can only be made using acoustic information rather than text alone. For instance, an inflection in vocal pitch at the end of an sentence can change its meaning from a statement to a question, thus requiring a question mark instead of a period. Secondly, cascading a series of post-processing steps into the speech recognition pipeline may lead to error propagation that hampers overall system performance \citep{knill18_impact, lu19_impact}. Finally, the pipeline system is evaluated for each post-processing component individually. This can result in individual components being optimised in isolation, at the expense of lower overall performance due to distribution shift \citep{sculley15-hidden-technical-debt}. As a result, post-processing can lead to systems failing to accurately predict orthographic transcriptions on datasets where it is required.

These issues and the need for dataset specific pre- or post-processing can be bypassed entirely by designing end-to-end models - from speech directly to orthographic transcripts \citep{graves14_towards, chan16_las}. E2E models have been shown to outperform traditional cascaded ASR systems, particularly when large amounts of labelled speech data is available \citep{hannun14_deepspeech, synnaeve20_endtoend, radford22_whipser}. What is more, E2E ASR systems require a single stage of evaluation; the ASR system is assessed on the cased and punctuated transcription outputs that are generated for the downstream application, giving a single, unified measure of overall performance. However, for the further development and refinement of these systems, it is important to have a benchmark targeting the specific challenges that end-to-end models face. 


\section{ESB Datasets}
\label{sec:datasets}
ESB comprises eight English speech recognition datasets, capturing a broad range of domains, acoustic conditions, speaker styles, and transcription requirements. We retain all punctuation, casing and formatting in the transcription outputs. Only annotation mistakes, such as double empty spaces, or annotation elements that cannot be considered transcriptions, such as \textit{$<$unk$>$}, are corrected. A comprehensive list of all transcription error corrections are detailed in Appendix~\ref{sub:appendix-error-correction}. As the objective of ESB is to motivate the development of end-to-end ASR, systems must use the same architecture across all datasets without any dataset-specific pre-processing or post-processing. Good performance requires systems capable of handling a range of audio and text conditions without any prior dataset-specific knowledge of the data distributions. The main datasets in ESB are accessible with permissive licensing. We also include three optional paid datasets that challenge interesting and unique domains of speech recognition, but do not require their inclusion for submission to the benchmark.
We describe the datasets below and in Table~\ref{tab:datasets-summary}, with additional details in Appendix~\ref{sec:appendix-data-preparation}.

\begin{table}[t]
\caption{Datasets description and statistics. Speaking style falls into one of three categories: narrated (N), oratory (O) and spontaneous (S). Datasets with multiple speaking styles are shown separated by a comma. Dataset sizes for the train/validation/test splits are quoted in hours of audio data. The transcription format is either normalised (Norm.), punctuated (P) or punctuated and cased (P+C).}
\begin{center}
\label{tab:datasets-summary}
\begin{tabular}{lllrl} 
\toprule
\textbf{Dataset} & \textbf{Domain} & \textbf{Style}        & \multicolumn{1}{l}{\textbf{Train/Val/Test}} & \textbf{Trans.}  \\ 
\midrule
LibriSpeech      & Audiobook                           & N & 960                                / 11                               / 11                                & Norm.                    \\
Common Voice              & Wikipedia                & N           & 1409                               / 27                               / 27                                & P+C                     \\
VoxPopuli        & EU Parliament                 & O           & 523                                / 5                                / 5                                 & P                    \\
TED-LIUM          & TED talks                           & O              & 454                                / 2                                / 3                                 & Norm.                    \\
GigaSpeech       & Audiobook, podcast, YouTube         & N, S           & 2500                               /12                               / 40                                & P                       \\
SPGISpeech       & Meetings                 & O, S           & 4900                               / 100                              / 100                               & P+C                     \\
Earnings-22      & Meetings                 & O, S               & 105                                / 5                                / 5                                 & P+C                     \\
AMI              & Meetings                            & S  & 78                                 / 9                                / 9                                 & P+C                     \\
SwitchBoard (optional)  & Telephone             & S               & 3572                               / 30                               / 7                                 & Norm.                    \\
CHiME-4 (optional) & Broadcast news & N & 19/11/7 & P+C \\
\bottomrule
\end{tabular}
\end{center}
\end{table}

\noindent
\textbf{LibriSpeech} \citep{panayotov15_libripseech} is a standard large-scale dataset for evaluating ASR systems. It consists of approximately 1000 hours of narrated audiobooks collected from the LibriVox\footnote{\url{https://librivox.org/}} project. Whilst instrumental in facilitating researchers to leverage a large body of pre-existing transcribed speech data, its standalone use presents limitations. The audiobook domain provides high-quality recording conditions that result in little to no background noise and the narrated speaking style lacks the acoustic and prosodic features of spontaneous speech. The transcriptions are non-orthographic without punctuation and casing. Since the books read are in the public domain, many contain antiquated language and writing styles atypical of modern-day speech.
We anticipate competitive systems to perform extremely well on LibriSpeech \citep{zhang20_pushing}. We include LibriSpeech in ESB to facilitate a comparison of performance between ideal speech recognition conditions and the more challenging settings presented by other datasets in the benchmark.
We use the standard split of train, validation (\emph{dev-clean}, \emph{dev-other}) and test sets (\emph{test-clean}, \emph{test-other}).

\noindent
\textbf{Common Voice} \citep{ardila20_commonvoice} is a series of crowd-sourced open-licensed speech datasets where speakers record text from Wikipedia in various languages. Since anyone can contribute recordings, there is significant variation in both audio quality and speakers. The audio conditions are challenging, with recording artefacts, accented speech, hesitations, and the presence of foreign words. The transcriptions are orthographic, with both casing and punctuation. However, the speaking style remains narrated (a shortcoming shared with LibriSpeech). We use the English subset of version 9.0 (27-4-2022), with approximately 1,400 hours and data splits provided therein.

\noindent
\textbf{VoxPopuli} \citep{wang21_voxpopuli} is a large-scale multilingual speech corpus consisting of data sourced from 2009-2020 European Parliament event recordings. Consequently, it occupies the unique domain of oratory, political speech, largely sourced from non-native speakers. We use the English subset with approximately 550 hours and the canonical data splits.

\noindent
\textbf{TED-LIUM} \citep{hernandez18_tedlium} is based on English-language TED Talk conference videos. 
The transcribed talks cover a range of different cultural, political, and academic topics, resulting in a technical vocabulary. We use Release 3 edition of the training set with approximately 450 hours and the legacy distribution of validation and test data, consistent with earlier releases for comparison.

\noindent
\textbf{GigaSpeech} \citep{chen21_gigaspeech} is a multi-domain English speech recognition corpus curated from audiobooks, podcasts and YouTube. It covers both narrated and spontaneous speech over a variety of topics, such as arts, science and sports. It is the only corpus in the benchmark to cover multiple domains. We use the large subset (2,500 hours) to train and the standard validation and test splits.

\noindent
\textbf{SPGISpeech} \citep{oneill21_kensho} is an English speech recognition corpus composed of company earnings calls that have been manually transcribed by S\&P Global, Inc. The transcriptions are fully-formatted according to a professional style guide for oratory and spontaneous speech. We train on the large subset (5,000 hours) and evaluate on the canonical validation and test splits.


\noindent
\textbf{Earnings-22} \citep{delrio22_earnings22} is a 119-hour corpus of English-language earnings calls collected from global companies. The dataset was developed with the goal of aggregating a broad range of speakers and accents covering a range of real-world financial topics. There is large diversity in the speakers and accents, with speakers taken from seven different language regions. To create train-validation-test splits, we partition the Earnings-22 corpus 90:5:5.

\noindent
\textbf{AMI} \citep{carletta07_ami, renals07_ami} comprises 100 hours of meeting recordings captured using different recording streams. The corpus contains manually annotated orthographic transcriptions of the meetings aligned at the word level. Individual samples of the AMI dataset contain very large audio files (between 10 and 60 minutes), which we segment to lengths feasible for training most ASR systems (for details, see Appendix~\ref{sec:appendix-data-preparation}).
We use the individual headset microphones (AMI-IHM) version of the dataset and the train, validation and test sets provided therein.

\noindent
\textbf{SwitchBoard (optional)} is a collection of two-sided conversational telephone speech amongst speakers from the US. Recorded over 10 years ago and at a lower sampling rate than the other corpora, it presents a noisy and challenging ASR problem. We partition 5\% of the SwitchBoard \citep{godfrey92_switchboard} corpus to form the validation split. We combine the remainder of the SwitchBoard corpus with Fisher \citep{ldc04_fisher_speech, ldc04_fisher_transcriptions} to form a train set consisting of approximately 3,600 hours. The test sets are the Hub5Eval2000 \citep{ldc02_hub5} data with two subsets: SwitchBoard and CallHome.

\noindent
\textbf{CHiME-4 (optional)} \citep{chime4_17} consists of narrated samples from the Wall Street Journal corpus \citep{ldc93_wsj}. Recordings are taken in challenging noisy environments using a 6-channel tablet based microphone array. We limit the official training data to single-channel and 18 hours by randomly selecting one of the six channels for each of the official training recordings. We use the official 1-channel development and test sets in their original annotated form.

SwitchBoard is a popular dataset for assessing ASR systems due to its unique telephone conversation domain. Alongside CHiME-4, these two datasets present challenging and noisy audio conditions. However, both datasets require payment for use. Thus, we include these corpora as optional extras in ESB; the score for these datasets is standalone and does not contribute to the overall benchmark score.

\section{Evaluation}
\paragraph{System Requirements}
ESB requires a single system to be defined and evaluated across the constituent datasets. The system must use the same architecture as well as training and evaluation algorithms for all datasets. This requirement includes using the same data pre- and post-processing of the audio inputs, target transcriptions, and system predictions. There is no restriction on the system being a single model, provided it is defined uniformly across all datasets. Given the range in size of the different datasets, hyper-parameter tuning is permitted, provided the algorithm for hyper-parameters tuning is consistent across datasets. The validation sets from each dataset are used to optimise system configurations and for hyper-parameter tuning, while the test sets are used only for the final evaluation.

Systems submitted to ESB may use any public or private data to train and develop their systems, including unlabelled audio data for pretraining, unlabelled text corpora for training language models (LMs) and labelled audio data for supervised training. However, systems may only use the ESB-distributed versions of the datasets included in the benchmark; in some cases, these datasets include different data preparation and train/validation/test splits than other public versions. In addition, systems may not use the unlabelled test data for training or development in any way, and may not share information across test samples in any way.


\paragraph{Metrics}
We evaluate system predictions against the target transcriptions using the word error rate (WER) metric. However, to encourage multi-domain systems capable of predicting orthographic transcriptions, we retain all dataset-specific transcription requirements (punctuation and casing) in our evaluation and evaluate systems on a per-dataset level. This decision leads to an \textit{orthographic WER} that is both punctuation \citep{kim01_prosody, KIM2003563} and case \citep{oneill21_kensho} sensitive. Punctuation symbols constitute their own words, such that incorrect punctuation is considered a word substitution error, missing punctuation a word deletion error and additional punctuation a word insertion error. To account for casing, we keep the upper and lower-case character sets distinct. Consequently, incorrect (resp. missing) capitalisation yields a word substitution (resp. deletion) error.

\paragraph{Benchmark Scoring}
We average WERs over individual datasets to give the final score. Through a macro-average, we aim to give a sense of aggregate system performance over all datasets. As with GLUE \citep{wang19_glue} and SuperGLUE \citep{wang19_superglue}, we lack a fair criterion with which to weigh the contribution of each dataset, and thus weigh each dataset equally. As LibriSpeech has multiple test sets (\textit{test-clean} and \textit{test-other}), we use an unweighted average of the WERs as the score for the dataset when computing the macro-average, so as not to weight it more heavily.


\paragraph{Leaderboard}
The ESB leaderboard keeps track of system submissions (similar to SemEval \citep{emerson22_semeval}, Kaggle\footnote{\url{https://www.kaggle.com}}, GLUE \citep{wang19_glue} and SuperGLUE \citep{wang19_superglue}). Data for the benchmark is available for download through Hugging Face Datasets \citep{lhoest21_datasets}. Each dataset contains standardised audio-transcriptions pairs for the training and validation sets. Only the unlablled audio samples are included for the test set. To submit a system, one must evaluate the system on the unlabelled audio test data for each of the ESB datasets and upload the predictions to \url{https://huggingface.co/spaces/esb/leaderboard} for scoring. The benchmark site details the orthographic WERs for the individual datasets and a macro-average of these scores to determine a system’s position on the leaderboard.

\section{Baselines}
We evaluate five different  systems. These baselines collectively represent current state-of-the-art approaches in E2E ASR. We describe them below, with additional details included in Appendix~\ref{sec:appendix-baseline}.


\noindent
\textbf{wav2vec 2.0 CTC} wav2vec 2.0 \citep{baevski20_wav2vec2} initialised from the official wav2vec 2.0 LARGE LV-60k checkpoint. The checkpoint is pretrained on an unsupervised task with 60k hours of unlabelled audio data from the LibriVox corpus. We follow \citet{baevski20_wav2vec2} and add a randomly initialised linear layer on top of the Transformer block to predict characters. The system is fine-tuned using the connectionist temporal classification (CTC) \citep{graves06_ctc} objective.

\noindent
\textbf{wav2vec 2.0 CTC + $n$-gram} wav2vec 2.0 CTC with a $5$-gram KenLM \citep{heafield11_kenlm} to perform LM boosted beam search decoding for CTC. The $5$-gram LM is trained on the train split transcriptions for each dataset.

\noindent
\textbf{wav2vec 2.0 AED} An attention-based encoder-decoder (AED) with a wav2vec 2.0 encoder and Transformer \citep{vaswani17_attention} decoder. Encoder weights are initialised with the wav2vec 2.0 LARGE checkpoint and decoder weights with the official BART LARGE \citep{lewis20_bart} checkpoint pretrained on 160 GB of text data. We follow \citet{li20-lna} and \citet{babu21-xlsr} in adding a randomly initialised adapter network to interface the encoder and decoder, consisting of three 1-dimensional CNN blocks. The system is fine-tuned using the cross-entropy objective.

\noindent
\textbf{Whisper AED} An AED network initialised with the encoder and decoder weights from the Whisper \citep{radford22_whipser} medium.en checkpoint pretrained on a supervised task with 680k hours of weakly labelled audio-transcription data. The system is fine-tuned using the cross-entropy objective.

\noindent
\textbf{Conformer RNN-T} The Conformer Transducer architecture \citep{gulati20_conformer}, combining a Conformer encoder with an RNN-Transducer (RNN-T) \citep{graves12_transducer} decoder. System weights from the NVIDIA NeMo \citep{kuchaiev19_nemo} XLARGE checkpoint\footnote{\url{https://catalog.ngc.nvidia.com/orgs/nvidia/teams/nemo/models/stt_en_conformer_transducer_xlarge}} trained on a supervised task with 24k hours of audio-transcription pairs. The system is fine-tuned using the RNN-T objective.

\section{Benchmark Results}
\begin{table}[t]
\caption{Baseline performance on the test sets and overall benchmark scores. We report orthographic WERs in \%. SwitchBoard, CallHome and CHiME-4 do not contribute to the benchmark score.}
\label{tab:baseline-results}
\begin{center}
\begin{tabular}{l>{\raggedleft\arraybackslash}p{1.25cm}>{\raggedleft\arraybackslash}p{1.25cm}>{\raggedleft\arraybackslash}p{1.25cm}>{\raggedleft\arraybackslash}p{1.25cm}>{\raggedleft\arraybackslash}p{1.25cm}}
\toprule
& \multicolumn{3}{c}{\textbf{wav2vec 2.0}} & \multicolumn{1}{c}{\textbf{Whisper}} &  \multicolumn{1}{c}{\textbf{Conformer}} \\
Dataset       & \multicolumn{1}{c}{\textbf{CTC}} & \multicolumn{1}{c}{$\substack{\textbf{CTC +}\\ \text{n-gram}}$} & \multicolumn{1}{c}{\textbf{AED}} & \multicolumn{1}{c}{\textbf{AED}} & \multicolumn{1}{c}{\textbf{RNN-T}}  \\ 
\midrule
LibriSpeech \textit{test-clean} & 2.9                              & 2.4                                                                                 & 2.8                                   & 2.2                                 & \textbf{2.0}                        \\
LibriSpeech \textit{test-other} & 7.5                              & 5.9                                                                                 & 5.8                                   & 5.2                                 & \textbf{4.0}                        \\
Common Voice                    & 26.1                             & 22.2                                                                                & 16.3                                  & 15.8                                & \textbf{14.8}                       \\
VoxPopuli                       & 11.4                             & 10.2                                                                                & 10.1                                  & 7.4                                 & \textbf{7.3}                        \\
TED-LIUM                        & 8.4                              & 6.7                                                                                 & 6.9                                   & \textbf{4.7}                        & 5.0                                 \\
GigaSpeech                      & 25.3                             & 22.0                                                                                & 23.4                                  & \textbf{17.3}                       & 18.6                                \\
SPGISpeech                      & 8.1                              & 7.1                                                                                 & \textbf{5.4}                          & 5.5                                 & 6.3                                 \\
Earnings-22                     & 26.0                             & 31.7                                                                                & 23.6                                  & \textbf{16.0}                       & 17.6                                \\
AMI                             & 32.0                             & 33.1                                                                                & 19.3                                  & \textbf{14.5}                       & 15.1                                \\ 
\midrule
SwitchBoard          & 16.1                             & 12.8                                                                                & 15.3                                  & \textbf{10.0}                       & 10.8                                \\
CallHome            & 26.6                             & 20.9                                                                                & 24.3                                  & \textbf{15.9}                       & 23.3                                \\
CHiME-4              & 29.2 & 	26.8 &	56.9 &	\textbf{12.7} &	14.2\\ 
\midrule
\textbf{ESB Score}                  & 17.8                             & 17.1                                                                                & 13.7                                  & \textbf{10.6}                       & 11.0                                \\
\bottomrule
\end{tabular}
\end{center}
\end{table}

We present performance on ESB for all baselines in Table~\ref{tab:baseline-results}. We quote orthographic WER for each dataset and a macro-average to yield to the overall benchmark score.

Amongst the wav2vec 2.0 baselines trained with unsupervised pretraining, CTC achieves competitive results on LibriSpeech \textit{test-clean}. Incorporating a LM with CTC + $n$-gram reduces the benchmark WER score by 0.7\% absolute, attaining significant gains on seven of the test sets. On Earnings-22 and AMI, CTC + $n$-gram performs worse than CTC. The AED architecture significantly outperforms CTC and CTC + $n$-gram on the datasets with transcription punctuation and character casing, namely Common Voice, SPGISpeech, Earnings-22 and AMI. It performs comparably to CTC + $n$-gram for the others. Overall, it achieves a score 3.4\% lower on ESB.

Whisper AED and Conformer RNN-T incorporate supervised pretraining. 
Whisper AED improves on wav2vec 2.0 AED for all datasets bar SPGISpeech, achieving the best performance on four of the nine ESB test sets and competitive scores on the others. Conformer RNN-T also performs strongly across the board, achieving the best performance on the remaining four ESB test sets. Whisper and RNN-T achieve comparable ESB scores, with average WERs of 10.6 and 11.0\% respectively.

A similar ranking pattern emerges on the optional datasets. CTC + $n$-gram improves on CTC for all three test sets. wav2vec 2.0 AED performs similarly to CTC and CTC + $n$-gram on SwitchBoard and CallHome, both test sets that are single-cased and un-punctuated. It fails on CHiME-4, the smallest of the ten training sets, where 18 hours of labelled audio data is unlikely enough to train this system. Whisper AED and Conformer RNN-T yield the strongest results overall, with Whisper the only model to achieve a WER significantly lower than 20\% on CallHome.

Notably, we find that results on LibriSpeech do not correlate with ESB score. The most performant results for the LibriSpeech test sets are the two lowest across all datasets, with 2.0 and 4.0\% respectively. The most competitive WERs for VoxPopuli, TED-LIUM and SPGISpeech all range between 4.7 and 10.0\%. Common Voice, GigaSpeech, AMI, SwitchBoard, CallHome and CHiME-4 all pose challenges even for the best performing systems, with WERs greater than 10.0\%. These results indicate that solving ESB is beyond the capabilities of current models and methods.

\begin{wraptable}{r}{0.35\textwidth}
\vspace{-22pt}
\caption{Minimum WERs for the baselines ranked highest to lowest with speaking style.}
\label{tab:baseline-results-with-style}
\centering
\begin{tabular}{lrl}
\toprule
\textbf{Dataset}       & \multicolumn{1}{c}{\textbf{Best}} & \textbf{Style}  \\ 
\midrule
GigaSpeech             & 17.3                                 & N, S    \\
Earnings-22            & 16.0                                 & S              \\
CallHome              & 15.9                                 & S              \\
Common Voice           & 14.8                                 & N                 \\
AMI                    & 14.5                                 & S              \\
CHiME-4                &  12.7                                & N              \\
SwitchBoard            & 10.0                                 & S              \\
VoxPopuli              & 7.3                                  & O                  \\
SPGISpeech             & 5.4                                  & O, S     \\
TED-LIUM               & 4.7                                  & O                  \\
LibriSpeech(o) & 4.0                                  & N                 \\
LibriSpeech(c) & 2.0                                  & N                 \\
\bottomrule
\end{tabular}
\vspace{-5pt}
\end{wraptable}

\section{Analysis}
Table~\ref{tab:baseline-results-with-style} displays the ranked WERs alongside the speaking style for the 12 test sets. Of the audio artefacts, the speaking style matters greatly and is reflected in the benchmark. Five of the seven test sets with the highest WERs contain some degree of spontaneous speech. The next three test sets all contain components of oratory speech. The two test sets with the lowest WERs contain narrated speech alone. Although Common Voice is narrated, it is an outlier and sits in fourth position. This is likely due to its crowd-sourced nature; the high variability in speakers, accents and quality pose difficulties for ASR systems. CHiME-4 is also narrated and ranks in sixth position. This is attributed to the fact it is only 18 hours, has high degrees of noise on the audio inputs and orthographic transcriptions. Whilst SPGISpeech contains oratory and spontaneous speech, it is similar in WER to the test sets that are oratory only (TED-LIUM and VoxPopuli). SPGISpeech potentially contains a much higher proportion of oratory speech than spontaneous and is the largest dataset in ESB (5,000 hours). 

\begin{wrapfigure}{r}{0.45\textwidth}
    \vspace{-19pt}
    \centering
    \includegraphics[width=0.45\textwidth, trim={0.5cm 1.17cm 0.5cm 0.75cm}]{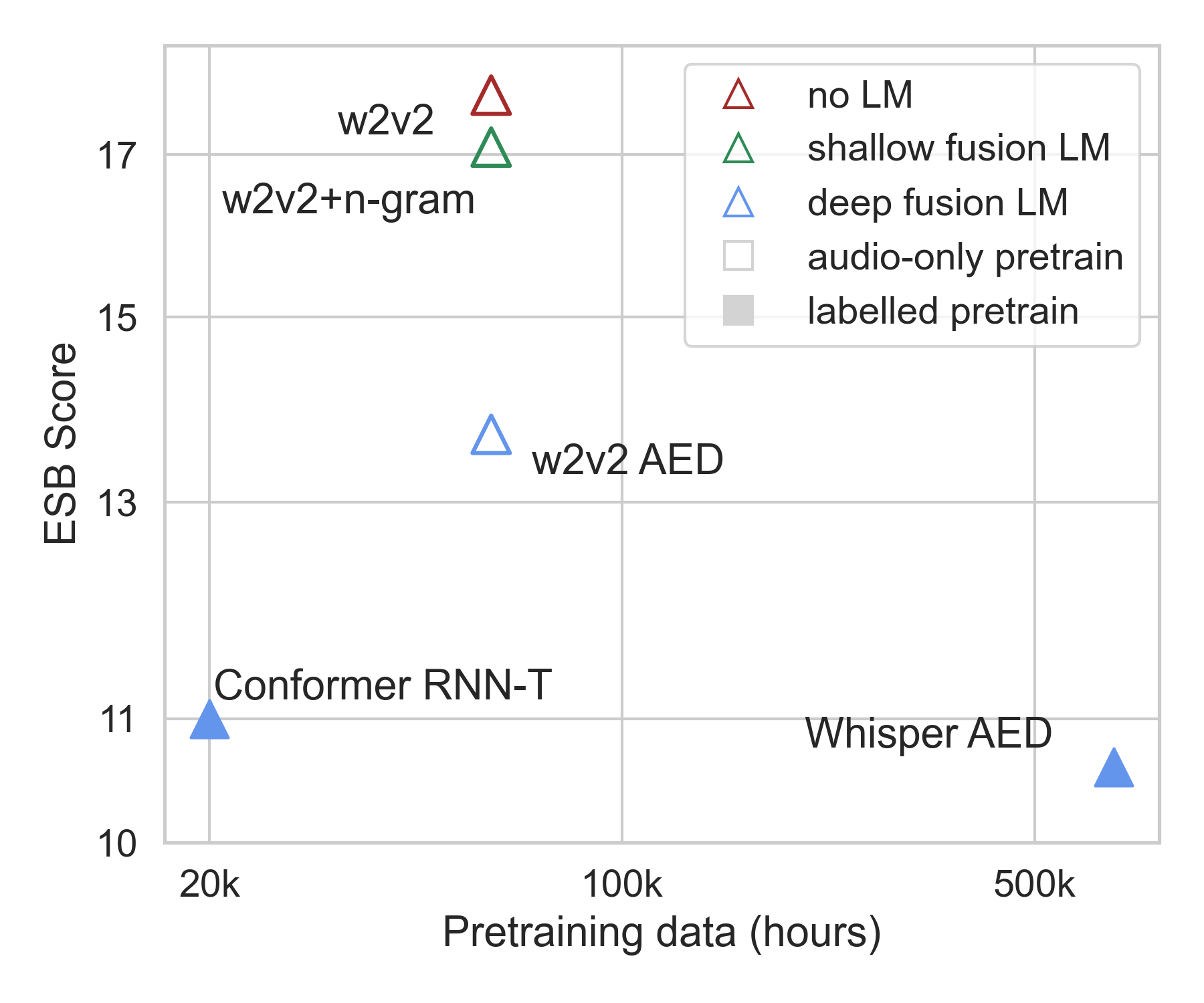}
    \caption{ESB score against pretraining data for the five baselines.}
    \label{fig:esc-vs-data}
\end{wrapfigure}

Figure~\ref{fig:esc-vs-data} plots ESB performance against pretraining data. wav2vec 2.0 is pretrained on an unsupervised task with 60k hours of unlabelled data from narrated audiobooks. The three wav2vec 2.0 based systems have the highest ESB score of the five baselines. Whisper AED and Conformer RNN-T are pretrained on 680k and 24k hours of labelled audio data from diverse sources, including multiple domains and speaking styles. These systems achieve competitive performance across the ESB test sets. This suggests that pretraining on diverse, labelled audio data facilitates ASR systems that can be applied to different datasets and domains.

To understand the impact of orthographic transcription features on system performance, we re-compute the per-dataset WERs by modifying outputs and targets: (i) remove punctuation, (ii) remove casing, (iii) apply full normalisation. We employ the full English text normaliser from \citet{radford22_whipser}, which removes filler words (``uh", ``uhm", ``mhm"), standardises number formatting (``0" to ``zero") and makes spellings consistent.

Table~\ref{tab:score-ablation} shows the orthographic ESB scores and the macro-average WERs (per-dataset scores are in Appendix~\ref{sec:appendix-score-ablation}). Removing punctuation yields a reduction of 2.0\% or more for all systems. Punctuation proves difficult for all five systems, but particularly so for CTC + $n$-gram which exhibits the largest reduction (4.1\%). The deep fusion systems have the lowest  absolute reductions in WER following punctuation removal (2.3, 2.0 and 2.3\% respectively). Casing further reduces the score by 0.5-0.6\%
for all systems. 
We observe another 0.5\% drop with full normalisation. In total, transcription features account for upwards of 3\% of the ESB score for all systems. This reveals the difficulty that orthographic transcription features pose for ASR systems.

\begin{table}[t]
\caption{The effect of punctuation, casing and full normalisation on benchmark score. We show the orthographic ESB score (no post-processing), score with punctuation removed, score with casing removed and score with full normalisation.}
\label{tab:score-ablation}
\begin{center}
\begin{tabular}{l>{\raggedleft\arraybackslash}p{1.25cm}>{\raggedleft\arraybackslash}p{1.25cm}>{\raggedleft\arraybackslash}p{1.25cm}>{\raggedleft\arraybackslash}p{1.25cm}>{\raggedleft\arraybackslash}p{1.25cm}}
\toprule
& \multicolumn{3}{c}{\textbf{wav2vec 2.0}} & \multicolumn{1}{c}{\textbf{Whisper}} &  \multicolumn{1}{c}{\textbf{Conformer}} \\
Score       & \multicolumn{1}{c}{\textbf{CTC}} & \multicolumn{1}{c}{$\substack{\text{\textbf{CTC +}}\\ \text{n-gram}}$} & \multicolumn{1}{c}{\textbf{AED}} & \multicolumn{1}{c}{\textbf{AED}} & \multicolumn{1}{c}{\textbf{RNN-T}}  \\ 
\midrule
Orthographic (ESB)             & 17.8                             & 17.1                                                                                & 13.7                                  & \textbf{10.6}                       & 11.0                                \\
- punctuation   & 14.8                             & 13.0                                                                                & 11.4                                  & \textbf{8.6}                        & 8.7                                 \\
- casing        & 14.3                             & 12.4                                                                                & 10.8                                  & \textbf{8.0}                        & 8.1                                 \\
- normalisation & 13.7                             & 11.6                                                                                & 10.3                                  & 7.4                                 & \textbf{7.2}                        \\
\bottomrule
\end{tabular}
\end{center}
\end{table}

To compare the E2E baselines with SoTA systems for individual datasets, we assess the baselines on comparable transcription conditions. Namely, we remove all transcription punctuation and character casing. For GigaSpeech and AMI, we also remove filler words. For Common Voice, we apply the full English text normaliser from \citet{radford22_whipser}. We list the SoTA results as of 09-2022, except for Earnings-22 where the entire dataset is used as a test-set only. Table~\ref{tab:baseline-results-with-sota} details the WERs under comparable conditions for the five E2E baselines and individual SoTA. The WER for the best forming E2E systems are to within 2.6\% of SoTA for the ESB test sets. The gap is wider for the optional datasets, standing at 7.6\% for CallHome. We achieve SoTA results on Common Voice, TED-LIUM and SPGISpeech with Conformer RNN-T, Whisper AED and wav2vec 2.0 AED respectively. These results demonstrate that E2E systems can be applied effectively to a range of different ASR datasets and domains, although there is still remaining future work on some datasets.

\begin{table}[t]
\caption{WER on the ESB test sets under non-orthographic conditions (case and punctuation normalised, removal of fillers for GigaSpeech and AMI, full normalisation for Common Voice). SoTA results are shown alongside those for the five baseline systems. The SoTA results for LibriSpeech are from \citet{zhang20_pushing}, Common Voice \citet{radford22_whipser}, VoxPopuli \citet{conneau22_xtreme_s}, TED-LIUM \citet{zhang22_bigssl}, GigaSpeech \citet{chen21_gigaspeech}, AMI \citet{zhang22_bigssl}, SwitchBoard and CallHome \citet{tuske21_limit}, and CHiME-4 \citet{du16_chime4}.}
\label{tab:baseline-results-with-sota}
\begin{center}
\begin{tabular}{l>{\raggedleft\arraybackslash}p{1cm}>{\raggedleft\arraybackslash}p{1cm}>{\raggedleft\arraybackslash}p{1cm}>{\raggedleft\arraybackslash}p{1cm}>{\raggedleft\arraybackslash}p{1cm}>{\raggedleft\arraybackslash}p{1.15cm}}
\toprule
& \multicolumn{3}{c}{\textbf{wav2vec 2.0}} & \multicolumn{1}{c}{\textbf{Whisper}} &  \multicolumn{1}{c}{\textbf{Conformer}}
\\
Dataset       & \multicolumn{1}{c}{\textbf{CTC}} & \multicolumn{1}{c}{$\substack{\text{\textbf{CTC +}}\\ \text{n-gram}}$} & \multicolumn{1}{c}{\textbf{AED}} & \multicolumn{1}{c}{\textbf{AED}} & \multicolumn{1}{c}{\textbf{Best E2E}} & \multicolumn{1}{c}{\textbf{SoTA}} \\ 
\midrule
LibriSpeech \textit{test-clean} & 2.9                              & 2.4                                                                                 & 2.8                                   & 2.2                                 & 2.0                       & 1.4                                \\
LibriSpeech \textit{test-other} & 7.5                              & 5.9                                                                                 & 5.8                                   & 5.2                                 & 4.0                       & 2.6                                \\
Common Voice                    & 21.9                             & 16.8                                                                                & 12.6                                  & 12.2                                & 9.7                       & 10.1                               \\
VoxPopuli                       & 10.3                             & 8.3                                                                                 & 9.8                                   & 7.2                                 & 7.1                       & 7.0                                \\
TED-LIUM                        & 8.4                              & 6.7                                                                                 & 6.9                                   & 4.7                        & 5.0                                & 5.0                                \\
GigaSpeech                      & 17.6                             & 14.3                                                                                & 14.3                                  & 10.5                       & 11.7                               & 10.5                               \\
SPGISpeech                      & 4.4                              & 3.3                                                                                 & 2.2                          & 2.4                                 & 2.7                                & 2.3                                \\
Earnings-22                     & 20.4                             & 20.5                                                                                & 19.6                                  & 11.5                       & 12.6                               & -                                  \\
AMI                             & 22.8                             & 20.6                                                                                & 15.2                                  & 10.4                       & 10.5                               & 7.8                                \\ 
\midrule
SwitchBoard                     & 14.1                             & 10.6                                                                                & 12.0                                  & 8.1                        & 8.8                                & 4.3                                \\
CallHome                       & 25.9                             & 20.2                                                                                & 19.8                                  & 14.4                       & 22.4                               & 6.8                                \\
CHiME-4 & 30.7 &	25.7 &	62.8 &	11.9 &	13.4 &	10.9 \\
\bottomrule
\end{tabular}
\end{center}
\end{table}





\section{Conclusion}

We introduce ESB, a benchmark for evaluating a end-to-end ASR systems across a broad range of speech domains. The eight datasets in our benchmark include speech recognition domains with different distributions for the audio artefacts and transcription requirements. We motivate the need for dataset-agnostic systems that can be applied across different domains without additional processing. We evaluate five different E2E systems on ESB, demonstrating how a single E2E system can be applied to different datasets and domains. In aggregate, systems pretrained on labelled data achieve better performance than those trained on unlabelled data, but still leave scope for improvement.

We believe that ESB offers a benchmark for developing improved speech recognition systems. Current results show that E2E systems near the performance of state-of-the-art systems on standard measures. Our analysis demonstrates the difficulty that different audio domains as well as punctuation and casing pose for ASR systems. As methods for pretraining ASR systems improve, we expect these issues to narrow as well. 




\subsubsection*{Reproducibility Statement}
ESB has been designed from the ground up to be accessible to everybody.
All mandatory datasets have permissive licences, are free to use, and can be downloaded with a single line of code via the Hugging Face Datasets library\footnote{ESB Datasets: \url{https://huggingface.co/datasets/esb/esb-datasets}}. SwitchBoard and CHiME-4 are optional, as we believe open-sourcing every core aspect of ESB is beneficial for promoting the research and development of ASR systems. In this regard, much work has gone into improving how the community can access the datasets included in ESB. Datasets exceeding 5,000 hours of training data were not considered for the benchmark, thus ensuring ESB remains practically feasible for academic research and that members of the speech community with modest computational requirements can submit to the benchmark.
Furthermore, we strongly encourage submissions to include all code, training logs, 
fine-tuned checkpoints, and evaluation runs necessary to reproduce a results.
All of our baseline checkpoints, training logs and code are fully open-sourced and can easily be accessed on the Hugging Face Hub\footnote{ESB Hub: \url{https://huggingface.co/esb}}.
Additionally, we describe all details regarding our baselines and evaluation strategies in Appendix~\ref{sec:appendix-baseline}.

\subsubsection*{Acknowledgments}
We thank Polina Kazakova and Anton Lozhkov for their help in preparing the ESB datasets for the Hugging Face Hub. We especially thank Ralf Schl\"{u}ter, Albert Zeyer, Georg Kucsko and Mark Gales for their in-depth feedback in the latter stages of this work (whilst we did not necessarily share the same views, we are certain that your feedback improved the paper and we are very grateful for this). We also thank Titouan Parcollet, Keenan Freyberg, Michael Schulman, Miguel del Rio, Quinten McNamara, Corey Miller, Guoguo Chen, Jiayu Du, Shinji Watanabe, Changhan Wang and EM Lewis-Jong for their comments on this work. We thank the individual dataset authors for their support in using the datasets for ESB. We gratefully acknowledge the support of Google's TPU Research Cloud (TRC) program in providing Cloud TPU resources for this research.

\bibliography{bibliography}

\begin{thebibliography}{81}
\providecommand{\natexlab}[1]{#1}
\providecommand{\url}[1]{\texttt{#1}}
\expandafter\ifx\csname urlstyle\endcsname\relax
  \providecommand{\doi}[1]{doi: #1}\else
  \providecommand{\doi}{doi: \begingroup \urlstyle{rm}\Url}\fi

\bibitem[Aks{\"e}nova et~al.(2021)Aks{\"e}nova, van Esch, Flynn, and
  Golik]{aksenova21_asrbenchmarks}
Al{\"e}na Aks{\"e}nova, Daan van Esch, James Flynn, and Pavel Golik.
\newblock {How Might We Create Better Benchmarks for Speech Recognition?}
\newblock In \emph{Proceedings of the 1st Workshop on Benchmarking: Past,
  Present and Future}, pp.\  22--34, Online, August 2021. Association for
  Computational Linguistics.
\newblock \doi{10.18653/v1/2021.bppf-1.4}.
\newblock URL \url{https://aclanthology.org/2021.bppf-1.4}.

\bibitem[Ardila et~al.(2020)Ardila, Branson, Davis, Kohler, Meyer, Henretty,
  Morais, Saunders, Tyers, and Weber]{ardila20_commonvoice}
Rosana Ardila, Megan Branson, Kelly Davis, Michael Kohler, Josh Meyer, Michael
  Henretty, Reuben Morais, Lindsay Saunders, Francis Tyers, and Gregor Weber.
\newblock {Common Voice: A Massively-Multilingual Speech Corpus}.
\newblock In \emph{Proceedings of the 12th Language Resources and Evaluation
  Conference}, pp.\  4218--4222, Marseille, France, May 2020. European Language
  Resources Association.
\newblock ISBN 979-10-95546-34-4.
\newblock URL \url{https://www.aclweb.org/anthology/2020.lrec-1.520}.

\bibitem[{Babu} et~al.(2021){Babu}, {Wang}, {Tjandra}, {Lakhotia}, {Xu},
  {Goyal}, {Singh}, {von Platen}, {Saraf}, {Pino}, {Baevski}, {Conneau}, and
  {Auli}]{babu21-xlsr}
Arun {Babu}, Changhan {Wang}, Andros {Tjandra}, Kushal {Lakhotia}, Qiantong
  {Xu}, Naman {Goyal}, Kritika {Singh}, Patrick {von Platen}, Yatharth {Saraf},
  Juan {Pino}, Alexei {Baevski}, Alexis {Conneau}, and Michael {Auli}.
\newblock {XLS-R: Self-supervised Cross-lingual Speech Representation Learning
  at Scale}.
\newblock \emph{arXiv e-prints}, art. arXiv:2111.09296, November 2021.

\bibitem[Babuschkin et~al.(2020)Babuschkin, Baumli, Bell, Bhupatiraju, Bruce,
  Buchlovsky, Budden, Cai, Clark, Danihelka, Fantacci, Godwin, Jones, Hemsley,
  Hennigan, Hessel, Hou, Kapturowski, Keck, Kemaev, King, Kunesch, Martens,
  Merzic, Mikulik, Norman, Quan, Papamakarios, Ring, Ruiz, Sanchez, Schneider,
  Sezener, Spencer, Srinivasan, Wang, Stokowiec, and Viola]{deepmind20_jax}
Igor Babuschkin, Kate Baumli, Alison Bell, Surya Bhupatiraju, Jake Bruce, Peter
  Buchlovsky, David Budden, Trevor Cai, Aidan Clark, Ivo Danihelka, Claudio
  Fantacci, Jonathan Godwin, Chris Jones, Ross Hemsley, Tom Hennigan, Matteo
  Hessel, Shaobo Hou, Steven Kapturowski, Thomas Keck, Iurii Kemaev, Michael
  King, Markus Kunesch, Lena Martens, Hamza Merzic, Vladimir Mikulik, Tamara
  Norman, John Quan, George Papamakarios, Roman Ring, Francisco Ruiz, Alvaro
  Sanchez, Rosalia Schneider, Eren Sezener, Stephen Spencer, Srivatsan
  Srinivasan, Luyu Wang, Wojciech Stokowiec, and Fabio Viola.
\newblock {The DeepMind JAX Ecosystem}, 2020.
\newblock URL \url{http://github.com/deepmind}.

\bibitem[Baevski et~al.(2020)Baevski, Zhou, Mohamed, and
  Auli]{baevski20_wav2vec2}
Alexei Baevski, Yuhao Zhou, Abdelrahman Mohamed, and Michael Auli.
\newblock {wav2vec 2.0: A Framework for Self-Supervised Learning of Speech
  Representations}.
\newblock In H.~Larochelle, M.~Ranzato, R.~Hadsell, M.F. Balcan, and H.~Lin
  (eds.), \emph{Advances in Neural Information Processing Systems}, volume~33,
  pp.\  12449--12460. Curran Associates, Inc., 2020.
\newblock URL
  \url{https://proceedings.neurips.cc/paper/2020/file/92d1e1eb1cd6f9fba3227870bb6d7f07-Paper.pdf}.

\bibitem[Beeferman et~al.(1998)Beeferman, Berger, and
  Lafferty]{beeferman98_cyberpunc}
Doug Beeferman, Adam Berger, and John Lafferty.
\newblock {Cyberpunc: A lightweight punctuation annotation system for speech}.
\newblock In \emph{In Proceedings of the IEEE International Conference on
  Acoustics, Speech and Signal Processing}, pp.\  689--692, 1998.

\bibitem[Carletta(2007)]{carletta07_ami}
Jean Carletta.
\newblock {Unleashing the killer corpus: experiences in creating the
  multi-everything AMI Meeting Corpus}.
\newblock \emph{Language Resources and Evaluation}, 41\penalty0 (2):\penalty0
  181--190, 2007.
\newblock ISSN 1574-020X.
\newblock \doi{10.1007/s10579-007-9040-x}.

\bibitem[Chan et~al.(2016)Chan, Jaitly, Le, and Vinyals]{chan16_las}
William Chan, Navdeep Jaitly, Quoc Le, and Oriol Vinyals.
\newblock {Listen, attend and spell: A neural network for large vocabulary
  conversational speech recognition}.
\newblock In \emph{2016 IEEE International Conference on Acoustics, Speech and
  Signal Processing (ICASSP)}, pp.\  4960--4964, 2016.
\newblock \doi{10.1109/ICASSP.2016.7472621}.

\bibitem[{Chan} et~al.(2021){Chan}, {Park}, {Lee}, {Zhang}, {Le}, and
  {Norouzi}]{chan21_speechstew}
William {Chan}, Daniel {Park}, Chris {Lee}, Yu~{Zhang}, Quoc {Le}, and Mohammad
  {Norouzi}.
\newblock {SpeechStew: Simply Mix All Available Speech Recognition Data to
  Train One Large Neural Network}.
\newblock \emph{arXiv e-prints}, art. arXiv:2104.02133, April 2021.

\bibitem[Chen(1999)]{chen1999_speech}
C~Julian Chen.
\newblock Speech recognition with automatic punctuation.
\newblock In \emph{Sixth European Conference on Speech Communication and
  Technology}, 1999.

\bibitem[{Chen} et~al.(2021){Chen}, {Chai}, {Wang}, {Du}, {Zhang}, {Weng},
  {Su}, {Povey}, {Trmal}, {Zhang}, {Jin}, {Khudanpur}, {Watanabe}, {Zhao},
  {Zou}, {Li}, {Yao}, {Wang}, {Wang}, {You}, and {Yan}]{chen21_gigaspeech}
Guoguo {Chen}, Shuzhou {Chai}, Guanbo {Wang}, Jiayu {Du}, Wei-Qiang {Zhang},
  Chao {Weng}, Dan {Su}, Daniel {Povey}, Jan {Trmal}, Junbo {Zhang}, Mingjie
  {Jin}, Sanjeev {Khudanpur}, Shinji {Watanabe}, Shuaijiang {Zhao}, Wei {Zou},
  Xiangang {Li}, Xuchen {Yao}, Yongqing {Wang}, Yujun {Wang}, Zhao {You}, and
  Zhiyong {Yan}.
\newblock {GigaSpeech: An Evolving, Multi-domain ASR Corpus with 10,000 Hours
  of Transcribed Audio}.
\newblock \emph{arXiv e-prints}, art. arXiv:2106.06909, June 2021.

\bibitem[Choquette et~al.(2021)Choquette, Gandhi, Giroux, Stam, and
  Krashinsky]{choquette21_a100}
Jack Choquette, Wishwesh Gandhi, Olivier Giroux, Nick Stam, and Ronny
  Krashinsky.
\newblock {NVIDIA A100 Tensor Core GPU: Performance and Innovation}.
\newblock \emph{IEEE Micro}, 41\penalty0 (2):\penalty0 29--35, 2021.
\newblock \doi{10.1109/MM.2021.3061394}.

\bibitem[Cieri et~al.(2004{\natexlab{a}})]{ldc04_fisher_speech}
Christopher Cieri et~al.
\newblock {Fisher English Training Speech Part 1 Speech LDC2004S13. Web
  Download}.
\newblock \emph{Linguistic Data Consortium}, 2004{\natexlab{a}}.

\bibitem[Cieri et~al.(2004{\natexlab{b}})]{ldc04_fisher_transcriptions}
Christopher Cieri et~al.
\newblock {Fisher English Training Speech Part 1 Transcripts LDC2004T19. Web
  Download}.
\newblock \emph{Linguistic Data Consortium}, 2004{\natexlab{b}}.

\bibitem[Cieri et~al.(2005{\natexlab{a}})]{ldc04_fisher_speech_2}
Christopher Cieri et~al.
\newblock {Fisher English Training Speech Part 2 Speech LDC2005S13. Web
  Download}.
\newblock \emph{Linguistic Data Consortium}, 2005{\natexlab{a}}.

\bibitem[Cieri et~al.(2005{\natexlab{b}})]{ldc04_fisher_transcriptions_2}
Christopher Cieri et~al.
\newblock {Fisher English Training Speech Part 2 Transcripts LDC2005T19. Web
  Download}.
\newblock \emph{Linguistic Data Consortium}, 2005{\natexlab{b}}.

\bibitem[{Conneau} et~al.(2022){Conneau}, {Bapna}, {Zhang}, {Ma}, {von Platen},
  {Lozhkov}, {Cherry}, {Jia}, {Rivera}, {Kale}, {Van Esch}, {Axelrod},
  {Khanuja}, {Clark}, {Firat}, {Auli}, {Ruder}, {Riesa}, and
  {Johnson}]{conneau22_xtreme_s}
Alexis {Conneau}, Ankur {Bapna}, Yu~{Zhang}, Min {Ma}, Patrick {von Platen},
  Anton {Lozhkov}, Colin {Cherry}, Ye~{Jia}, Clara {Rivera}, Mihir {Kale}, Daan
  {Van Esch}, Vera {Axelrod}, Simran {Khanuja}, Jonathan~H. {Clark}, Orhan
  {Firat}, Michael {Auli}, Sebastian {Ruder}, Jason {Riesa}, and Melvin
  {Johnson}.
\newblock {XTREME-S: Evaluating Cross-lingual Speech Representations}.
\newblock \emph{arXiv e-prints}, art. arXiv:2203.10752, March 2022.

\bibitem[{Del Rio} et~al.(2022){Del Rio}, {Ha}, {McNamara}, {Miller}, and
  {Chandra}]{delrio22_earnings22}
Miguel {Del Rio}, Peter {Ha}, Quinten {McNamara}, Corey {Miller}, and Shipra
  {Chandra}.
\newblock {Earnings-22: A Practical Benchmark for Accents in the Wild}.
\newblock \emph{arXiv e-prints}, art. arXiv:2203.15591, March 2022.

\bibitem[Du et~al.(2016)Du, Tu, Sun, Ma, kun Wang, Pan, Liu, Chen, and
  Lee]{du16_chime4}
Jun Du, Yan-Hui Tu, Lei Sun, Feng Ma, Hai kun Wang, Jia Pan, Cong Liu,
  Jing-Dong Chen, and Chin-Hui Lee.
\newblock {The USTC-iFlytek system for CHiME-4 challenge}.
\newblock \emph{4th International Workshop on Speech Processing in Everyday
  Environments}, 2016.

\bibitem[Emerson et~al.(2022)Emerson, Schluter, Stanovsky, Kumar, Palmer,
  Schneider, Singh, and Ratan]{emerson22_semeval}
Guy Emerson, Natalie Schluter, Gabriel Stanovsky, Ritesh Kumar, Alexis Palmer,
  Nathan Schneider, Siddharth Singh, and Shyam Ratan (eds.).
\newblock \emph{Proceedings of the 16th International Workshop on Semantic
  Evaluation (SemEval-2022)}, Seattle, United States, July 2022. Association
  for Computational Linguistics.
\newblock URL \url{https://aclanthology.org/2022.semeval-1.0}.

\bibitem[{Galvez} et~al.(2021){Galvez}, {Diamos}, {Ciro}, {Cer{\'o}n},
  {Achorn}, {Gopi}, {Kanter}, {Lam}, {Mazumder}, and {Janapa
  Reddi}]{galvez21_peoples_speech}
Daniel {Galvez}, Greg {Diamos}, Juan {Ciro}, Juan~Felipe {Cer{\'o}n}, Keith
  {Achorn}, Anjali {Gopi}, David {Kanter}, Maximilian {Lam}, Mark {Mazumder},
  and Vijay {Janapa Reddi}.
\newblock {The People's Speech: A Large-Scale Diverse English Speech
  Recognition Dataset for Commercial Usage}.
\newblock \emph{arXiv e-prints}, art. arXiv:2111.09344, November 2021.

\bibitem[Garcés Díaz-Munío et~al.(2021)Garcés Díaz-Munío,
  Silvestre-Cerdà, Jorge, Giménez, Iranzo-Sánchez, Baquero-Arnal, Roselló,
  de~Martos, Civera, Sanchis, and Juan]{europarlasr2021}
Gonçal~V. Garcés Díaz-Munío, Joan~Albert Silvestre-Cerdà, Javier Jorge,
  Adrià Giménez, Javier Iranzo-Sánchez, Pau Baquero-Arnal, Nahuel Roselló,
  Alejandro Pérez-González de~Martos, Jorge Civera, Albert Sanchis, and
  Alfons Juan.
\newblock {Europarl-ASR: A Large Corpus of Parliamentary Debates for Streaming
  ASR Benchmarking and Speech Data Filtering/Verbatimization}.
\newblock In \emph{Proc. Interspeech 2021}, pp.\  3695--3699, Brno (Czech
  Republic), 2021.
\newblock \doi{10.21437/Interspeech.2021-1905}.

\bibitem[Garofolo et~al.(1993{\natexlab{a}})Garofolo, Lamel, Fisher, Fiscus,
  and Pallett]{garofolo93_darpa}
John~S Garofolo, Lori~F Lamel, William~M Fisher, Jonathan~G Fiscus, and David~S
  Pallett.
\newblock {DARPA TIMIT Acoustic-Phonetic Continuous Speech Corpus CD-ROM.}
\newblock \emph{NASA STI/Recon Technical Report}, 93:\penalty0 27403,
  1993{\natexlab{a}}.

\bibitem[Garofolo et~al.(1993{\natexlab{b}})]{ldc93_wsj}
John~S. Garofolo et~al.
\newblock {CSR-I (WSJ0) Complete LDC93S6A. Web Download}.
\newblock \emph{Linguistic Data Consortium}, 1993{\natexlab{b}}.

\bibitem[Godfrey et~al.(1992)Godfrey, Holliman, and
  McDaniel]{godfrey92_switchboard}
J.J. Godfrey, E.C. Holliman, and J.~McDaniel.
\newblock {SWITCHBOARD: telephone speech corpus for research and development}.
\newblock In \emph{[Proceedings] ICASSP-92: 1992 IEEE International Conference
  on Acoustics, Speech, and Signal Processing}, volume~1, pp.\  517--520 vol.1,
  1992.
\newblock \doi{10.1109/ICASSP.1992.225858}.

\bibitem[Gravano et~al.(2009)Gravano, Jansche, and
  Bacchiani]{gravano09_restoring}
Agustin Gravano, Martin Jansche, and Michiel Bacchiani.
\newblock Restoring punctuation and capitalization in transcribed speech.
\newblock In \emph{2009 IEEE International Conference on Acoustics, Speech and
  Signal Processing}, pp.\  4741--4744, 2009.
\newblock \doi{10.1109/ICASSP.2009.4960690}.

\bibitem[{Graves}(2012)]{graves12_transducer}
Alex {Graves}.
\newblock {Sequence Transduction with Recurrent Neural Networks}.
\newblock \emph{arXiv e-prints}, art. arXiv:1211.3711, November 2012.

\bibitem[Graves \& Jaitly(2014)Graves and Jaitly]{graves14_towards}
Alex Graves and Navdeep Jaitly.
\newblock {Towards End-to-End Speech Recognition with Recurrent Neural
  Networks}.
\newblock In \emph{Proceedings of the 31st International Conference on
  International Conference on Machine Learning - Volume 32}, ICML'14, pp.\
  II–1764–II–1772. JMLR.org, 2014.

\bibitem[Graves et~al.(2006)Graves, Fern\'{a}ndez, Gomez, and
  Schmidhuber]{graves06_ctc}
Alex Graves, Santiago Fern\'{a}ndez, Faustino Gomez, and J\"{u}rgen
  Schmidhuber.
\newblock {Connectionist Temporal Classification: Labelling Unsegmented
  Sequence Data with Recurrent Neural Networks}.
\newblock In \emph{Proceedings of the 23rd International Conference on Machine
  Learning}, ICML '06, pp.\  369–376, New York, NY, USA, 2006. Association
  for Computing Machinery.
\newblock ISBN 1595933832.
\newblock \doi{10.1145/1143844.1143891}.
\newblock URL \url{https://doi.org/10.1145/1143844.1143891}.

\bibitem[{Gulati} et~al.(2020){Gulati}, {Qin}, {Chiu}, {Parmar}, {Zhang}, {Yu},
  {Han}, {Wang}, {Zhang}, {Wu}, and {Pang}]{gulati20_conformer}
Anmol {Gulati}, James {Qin}, Chung-Cheng {Chiu}, Niki {Parmar}, Yu~{Zhang},
  Jiahui {Yu}, Wei {Han}, Shibo {Wang}, Zhengdong {Zhang}, Yonghui {Wu}, and
  Ruoming {Pang}.
\newblock {Conformer: Convolution-augmented Transformer for Speech
  Recognition}.
\newblock \emph{arXiv e-prints}, art. arXiv:2005.08100, May 2020.

\bibitem[{Hannun} et~al.(2014){Hannun}, {Case}, {Casper}, {Catanzaro},
  {Diamos}, {Elsen}, {Prenger}, {Satheesh}, {Sengupta}, {Coates}, and
  {Ng}]{hannun14_deepspeech}
Awni {Hannun}, Carl {Case}, Jared {Casper}, Bryan {Catanzaro}, Greg {Diamos},
  Erich {Elsen}, Ryan {Prenger}, Sanjeev {Satheesh}, Shubho {Sengupta}, Adam
  {Coates}, and Andrew~Y. {Ng}.
\newblock {Deep Speech: Scaling up end-to-end speech recognition}.
\newblock \emph{arXiv e-prints}, art. arXiv:1412.5567, December 2014.

\bibitem[Heafield(2011)]{heafield11_kenlm}
Kenneth Heafield.
\newblock {KenLM: Faster and Smaller Language Model Queries}.
\newblock In \emph{Proceedings of the Sixth Workshop on Statistical Machine
  Translation}, pp.\  187--197, Edinburgh, Scotland, July 2011. Association for
  Computational Linguistics.
\newblock URL \url{https://aclanthology.org/W11-2123}.

\bibitem[Heafield et~al.(2013)Heafield, Pouzyrevsky, Clark, and
  Koehn]{heafield13_scalable}
Kenneth Heafield, Ivan Pouzyrevsky, Jonathan~H. Clark, and Philipp Koehn.
\newblock {Scalable Modified {K}neser-{N}ey Language Model Estimation}.
\newblock In \emph{Proceedings of the 51st Annual Meeting of the Association
  for Computational Linguistics (Volume 2: Short Papers)}, pp.\  690--696,
  Sofia, Bulgaria, August 2013. Association for Computational Linguistics.
\newblock URL \url{https://www.aclweb.org/anthology/P13-2121}.

\bibitem[Heek et~al.(2020)Heek, Levskaya, Oliver, Ritter, Rondepierre, Steiner,
  and van {Z}ee]{heek20_flax}
Jonathan Heek, Anselm Levskaya, Avital Oliver, Marvin Ritter, Bertrand
  Rondepierre, Andreas Steiner, and Marc van {Z}ee.
\newblock {F}lax: A neural network library and ecosystem for {JAX}, 2020.
\newblock URL \url{http://github.com/google/flax}.

\bibitem[Hernandez et~al.(2018)Hernandez, Nguyen, Ghannay, Tomashenko, and
  Est{\`e}ve]{hernandez18_tedlium}
Fran{\c{c}}ois Hernandez, Vincent Nguyen, Sahar Ghannay, Natalia Tomashenko,
  and Yannick Est{\`e}ve.
\newblock {TED-LIUM 3: Twice as Much Data and Corpus Repartition for
  Experiments on Speaker Adaptation}.
\newblock In \emph{Speech and Computer}, pp.\  198--208. Springer International
  Publishing, 2018.

\bibitem[Howard \& Ruder(2018)Howard and Ruder]{howard18_universal}
Jeremy Howard and Sebastian Ruder.
\newblock Universal language model fine-tuning for text classification.
\newblock In \emph{Proceedings of the 56th Annual Meeting of the Association
  for Computational Linguistics (Volume 1: Long Papers)}, pp.\  328--339,
  Melbourne, Australia, July 2018. Association for Computational Linguistics.
\newblock \doi{10.18653/v1/P18-1031}.
\newblock URL \url{https://aclanthology.org/P18-1031}.

\bibitem[Jouppi et~al.(2020)Jouppi, Yoon, Kurian, Li, Patil, Laudon, Young, and
  Patterson]{jouppi20_tpu}
Norman~P. Jouppi, Doe~Hyun Yoon, George Kurian, Sheng Li, Nishant Patil, James
  Laudon, Cliff Young, and David Patterson.
\newblock {A Domain-Specific Supercomputer for Training Deep Neural Networks}.
\newblock \emph{Commun. ACM}, 63\penalty0 (7):\penalty0 67–78, jun 2020.
\newblock ISSN 0001-0782.
\newblock \doi{10.1145/3360307}.
\newblock URL \url{https://doi.org/10.1145/3360307}.

\bibitem[Kim \& Stern(2008)Kim and Stern]{wada_19}
Chanwoo Kim and Richard Stern.
\newblock {Robust signal-to-noise ratio estimation based on waveform amplitude
  distribution analysis}.
\newblock pp.\  2598--2601, 09 2008.
\newblock \doi{10.21437/Interspeech.2008-644}.

\bibitem[Kim \& Stern(2016)Kim and Stern]{kim16_pncc}
Chanwoo Kim and Richard~M. Stern.
\newblock {Power-Normalized Cepstral Coefficients (PNCC) for Robust Speech
  Recognition}.
\newblock \emph{IEEE/ACM Transactions on Audio, Speech, and Language
  Processing}, 24\penalty0 (7):\penalty0 1315--1329, 2016.
\newblock \doi{10.1109/TASLP.2016.2545928}.

\bibitem[Kim \& Woodland(2001)Kim and Woodland]{kim01_prosody}
Ji-Hwan Kim and P.~C. Woodland.
\newblock {The use of prosody in a combined system for punctuation generation
  and speech recognition}.
\newblock In \emph{Proc. 7th European Conference on Speech Communication and
  Technology (Eurospeech 2001)}, pp.\  2757--2760, 2001.
\newblock \doi{10.21437/Eurospeech.2001-645}.

\bibitem[Kim \& Woodland(2003)Kim and Woodland]{KIM2003563}
Ji-Hwan Kim and Philip~C Woodland.
\newblock A combined punctuation generation and speech recognition system and
  its performance enhancement using prosody.
\newblock \emph{Speech Communication}, 41\penalty0 (4):\penalty0 563--577,
  2003.
\newblock ISSN 0167-6393.
\newblock \doi{https://doi.org/10.1016/S0167-6393(03)00049-9}.
\newblock URL
  \url{https://www.sciencedirect.com/science/article/pii/S0167639303000499}.

\bibitem[Kingma \& Ba(2015)Kingma and Ba]{diederik15_adam}
Diederik~P. Kingma and Jimmy Ba.
\newblock {Adam: A Method for Stochastic Optimization}.
\newblock In \emph{ICLR (Poster)}, 2015.
\newblock URL \url{http://arxiv.org/abs/1412.6980}.

\bibitem[Knill et~al.(2018)Knill, Gales, Kyriakopoulos, Malinin, Ragni, Wang,
  and Caines]{knill18_impact}
Kate Knill, Mark Gales, Konstantinos Kyriakopoulos, Andrey Malinin, Anton
  Ragni, Yu~Wang, and Andrew Caines.
\newblock {Impact of ASR performance on free speaking language assessment}.
\newblock In \emph{Proc. Interspeech 2018}, pp.\  1641--1645. International
  Speech Communication Association (ISCA), 2018.

\bibitem[Koh et~al.(2019)Koh, Mislan, Khoo, Ang, Ang, Ng, and
  Tan]{national_speech_19}
Jia Koh, Aqilah Mislan, Kevin Khoo, Brian Ang, Wilson Ang, Charmaine Ng, and
  Ying-Ying Tan.
\newblock {Building the Singapore English National Speech Corpus}.
\newblock pp.\  321--325, 09 2019.
\newblock \doi{10.21437/Interspeech.2019-1525}.

\bibitem[{Kuchaiev} et~al.(2019){Kuchaiev}, {Li}, {Nguyen}, {Hrinchuk},
  {Leary}, {Ginsburg}, {Kriman}, {Beliaev}, {Lavrukhin}, {Cook}, {Castonguay},
  {Popova}, {Huang}, and {Cohen}]{kuchaiev19_nemo}
Oleksii {Kuchaiev}, Jason {Li}, Huyen {Nguyen}, Oleksii {Hrinchuk}, Ryan
  {Leary}, Boris {Ginsburg}, Samuel {Kriman}, Stanislav {Beliaev}, Vitaly
  {Lavrukhin}, Jack {Cook}, Patrice {Castonguay}, Mariya {Popova}, Jocelyn
  {Huang}, and Jonathan~M. {Cohen}.
\newblock {NeMo: a toolkit for building AI applications using Neural Modules}.
\newblock \emph{arXiv e-prints}, art. arXiv:1909.09577, September 2019.

\bibitem[Kudo \& Richardson(2018)Kudo and Richardson]{kudo18_sentencepiece}
Taku Kudo and John Richardson.
\newblock {{S}entence{P}iece: A simple and language independent subword
  tokenizer and detokenizer for Neural Text Processing}.
\newblock In \emph{Proceedings of the 2018 Conference on Empirical Methods in
  Natural Language Processing: System Demonstrations}, pp.\  66--71, Brussels,
  Belgium, November 2018. Association for Computational Linguistics.
\newblock \doi{10.18653/v1/D18-2012}.
\newblock URL \url{https://aclanthology.org/D18-2012}.

\bibitem[Lewis et~al.(2020)Lewis, Liu, Goyal, Ghazvininejad, Mohamed, Levy,
  Stoyanov, and Zettlemoyer]{lewis20_bart}
Mike Lewis, Yinhan Liu, Naman Goyal, Marjan Ghazvininejad, Abdelrahman Mohamed,
  Omer Levy, Veselin Stoyanov, and Luke Zettlemoyer.
\newblock {BART: Denoising Sequence-to-Sequence Pre-training for Natural
  Language Generation, Translation, and Comprehension}.
\newblock In \emph{Proceedings of the 58th Annual Meeting of the Association
  for Computational Linguistics}, pp.\  7871--7880, Online, July 2020.
  Association for Computational Linguistics.
\newblock \doi{10.18653/v1/2020.acl-main.703}.
\newblock URL \url{https://aclanthology.org/2020.acl-main.703}.

\bibitem[Lhoest et~al.(2021)Lhoest, Villanova~del Moral, Jernite, Thakur, von
  Platen, Patil, Chaumond, Drame, Plu, Tunstall, Davison, {\v{S}}a{\v{s}}ko,
  Chhablani, Malik, Brandeis, Le~Scao, Sanh, Xu, Patry, McMillan-Major, Schmid,
  Gugger, Delangue, Matussi{\`e}re, Debut, Bekman, Cistac, Goehringer, Mustar,
  Lagunas, Rush, and Wolf]{lhoest21_datasets}
Quentin Lhoest, Albert Villanova~del Moral, Yacine Jernite, Abhishek Thakur,
  Patrick von Platen, Suraj Patil, Julien Chaumond, Mariama Drame, Julien Plu,
  Lewis Tunstall, Joe Davison, Mario {\v{S}}a{\v{s}}ko, Gunjan Chhablani,
  Bhavitvya Malik, Simon Brandeis, Teven Le~Scao, Victor Sanh, Canwen Xu,
  Nicolas Patry, Angelina McMillan-Major, Philipp Schmid, Sylvain Gugger,
  Cl{\'e}ment Delangue, Th{\'e}o Matussi{\`e}re, Lysandre Debut, Stas Bekman,
  Pierric Cistac, Thibault Goehringer, Victor Mustar, Fran{\c{c}}ois Lagunas,
  Alexander Rush, and Thomas Wolf.
\newblock {Datasets: A Community Library for Natural Language Processing}.
\newblock In \emph{Proceedings of the 2021 Conference on Empirical Methods in
  Natural Language Processing: System Demonstrations}, pp.\  175--184, Online
  and Punta Cana, Dominican Republic, November 2021. Association for
  Computational Linguistics.
\newblock URL \url{https://aclanthology.org/2021.emnlp-demo.21}.

\bibitem[{Li} et~al.(2020){Li}, {Wang}, {Tang}, {Tran}, {Tang}, {Pino},
  {Baevski}, {Conneau}, and {Auli}]{li20-lna}
Xian {Li}, Changhan {Wang}, Yun {Tang}, Chau {Tran}, Yuqing {Tang}, Juan
  {Pino}, Alexei {Baevski}, Alexis {Conneau}, and Michael {Auli}.
\newblock {Multilingual Speech Translation with Efficient Finetuning of
  Pretrained Models}.
\newblock \emph{arXiv e-prints}, art. arXiv:2010.12829, October 2020.

\bibitem[{Likhomanenko} et~al.(2020){Likhomanenko}, {Xu}, {Pratap},
  {Tomasello}, {Kahn}, {Avidov}, {Collobert}, and
  {Synnaeve}]{likhomanenko20_arxiv}
Tatiana {Likhomanenko}, Qiantong {Xu}, Vineel {Pratap}, Paden {Tomasello},
  Jacob {Kahn}, Gilad {Avidov}, Ronan {Collobert}, and Gabriel {Synnaeve}.
\newblock {Rethinking Evaluation in ASR: Are Our Models Robust Enough?}
\newblock \emph{arXiv e-prints}, art. arXiv:2010.11745, October 2020.

\bibitem[{Linguistic Data Consortium}(2002)]{ldc02_hub5}
{Linguistic Data Consortium}.
\newblock {2000 HUB5 English Evaluation Transcripts LDC2002T43. Web Download}.
\newblock \emph{Linguistic Data Consortium}, 2002.

\bibitem[Lita et~al.(2003)Lita, Ittycheriah, Roukos, and
  Kambhatla]{lita_03_truecaseing}
Lucian~Vlad Lita, Abe Ittycheriah, Salim Roukos, and Nanda Kambhatla.
\newblock {TRuEcasIng}.
\newblock In \emph{Proceedings of the 41st Annual Meeting on Association for
  Computational Linguistics - Volume 1}, ACL '03, pp.\  152–159, USA, 2003.
  Association for Computational Linguistics.
\newblock \doi{10.3115/1075096.1075116}.
\newblock URL \url{https://doi.org/10.3115/1075096.1075116}.

\bibitem[Lu et~al.(2019)Lu, Gales, Knill, Manakul, Wang, and Wang]{lu19_impact}
Y.~Lu, Mark~J.F. Gales, Kate~M. Knill, P.~Manakul, L.~Wang, and Y.~Wang.
\newblock {Impact of ASR Performance on Spoken Grammatical Error Detection}.
\newblock In \emph{Proc. Interspeech 2019}, pp.\  1876--1880. International
  Speech Communication Association (ISCA), 2019.
\newblock \doi{10.21437/Interspeech.2019-1706}.

\bibitem[McIntosh \& {Cambridge University Press.}(2015)McIntosh and {Cambridge
  University Press.}]{mcintosh15_dictionary}
Colin McIntosh and {Cambridge University Press.}
\newblock \emph{{Cambridge Advanced Learner's Dictionary}}.
\newblock Cambridge University Press, Cambridge, England, 4 edition, 2015.

\bibitem[Ney et~al.(1994)Ney, Essen, and Kneser]{ney94_kneser_ney}
Hermann Ney, Ute Essen, and Reinhard Kneser.
\newblock On structuring probabilistic dependences in stochastic language
  modelling.
\newblock \emph{Computer Speech \& Language}, 8\penalty0 (1):\penalty0 1--38,
  1994.
\newblock ISSN 0885-2308.
\newblock \doi{https://doi.org/10.1006/csla.1994.1001}.
\newblock URL
  \url{https://www.sciencedirect.com/science/article/pii/S0885230884710011}.

\bibitem[NIST(1998)]{nist98_snor}
{National Institute of Standards and Technology} NIST.
\newblock {The 1998 HUB-4 Evaluation Plan for recognition of Broadcast News, in
  English}, 1998.
\newblock URL
  \url{https://mig.nist.gov/MIG_Website/tests/bnr/1998/hub4e_98_spec.html}.

\bibitem[O’Neill et~al.(2021)O’Neill, Lavrukhin, Majumdar, Noroozi, Zhang,
  Kuchaiev, Balam, Dovzhenko, Freyberg, Shulman, Ginsburg, Watanabe, and
  Kucsko]{oneill21_kensho}
Patrick~K. O’Neill, Vitaly Lavrukhin, Somshubra Majumdar, Vahid Noroozi,
  Yuekai Zhang, Oleksii Kuchaiev, Jagadeesh Balam, Yuliya Dovzhenko, Keenan
  Freyberg, Michael~D. Shulman, Boris Ginsburg, Shinji Watanabe, and Georg
  Kucsko.
\newblock {SPGISpeech: 5,000 Hours of Transcribed Financial Audio for Fully
  Formatted End-to-End Speech Recognition}.
\newblock In \emph{Proc. Interspeech 2021}, pp.\  1434--1438, 2021.
\newblock \doi{10.21437/Interspeech.2021-1860}.

\bibitem[Panayotov et~al.(2015)Panayotov, Chen, Povey, and
  Khudanpur]{panayotov15_libripseech}
Vassil Panayotov, Guoguo Chen, Daniel Povey, and Sanjeev Khudanpur.
\newblock {Librispeech: An ASR corpus based on public domain audio books}.
\newblock In \emph{2015 IEEE International Conference on Acoustics, Speech and
  Signal Processing (ICASSP)}, pp.\  5206--5210, 2015.
\newblock \doi{10.1109/ICASSP.2015.7178964}.

\bibitem[{Park} et~al.(2019){Park}, {Chan}, {Zhang}, {Chiu}, {Zoph}, {Cubuk},
  and {Le}]{park19_specaug}
Daniel~S. {Park}, William {Chan}, Yu~{Zhang}, Chung-Cheng {Chiu}, Barret
  {Zoph}, Ekin~D. {Cubuk}, and Quoc~V. {Le}.
\newblock {SpecAugment: A Simple Data Augmentation Method for Automatic Speech
  Recognition}.
\newblock \emph{arXiv e-prints}, art. arXiv:1904.08779, April 2019.

\bibitem[Paszke et~al.(2019)Paszke, Gross, Massa, Lerer, Bradbury, Chanan,
  Killeen, Lin, Gimelshein, Antiga, Desmaison, Kopf, Yang, DeVito, Raison,
  Tejani, Chilamkurthy, Steiner, Fang, Bai, and Chintala]{paszke19_pytorch}
Adam Paszke, Sam Gross, Francisco Massa, Adam Lerer, James Bradbury, Gregory
  Chanan, Trevor Killeen, Zeming Lin, Natalia Gimelshein, Luca Antiga, Alban
  Desmaison, Andreas Kopf, Edward Yang, Zachary DeVito, Martin Raison, Alykhan
  Tejani, Sasank Chilamkurthy, Benoit Steiner, Lu~Fang, Junjie Bai, and Soumith
  Chintala.
\newblock {PyTorch: An Imperative Style, High-Performance Deep Learning
  Library}.
\newblock In H.~Wallach, H.~Larochelle, A.~Beygelzimer, F.~d\textquotesingle
  Alch\'{e}-Buc, E.~Fox, and R.~Garnett (eds.), \emph{Advances in Neural
  Information Processing Systems 32}, pp.\  8024--8035. Curran Associates,
  Inc., 2019.
\newblock URL
  \url{http://papers.neurips.cc/paper/9015-pytorch-an-imperative-style-high-performance-deep-learning-library.pdf}.

\bibitem[Povey et~al.(2011)Povey, Ghoshal, Boulianne, Burget, Glembek, Goel,
  Hannemann, Motlicek, Qian, Schwarz, Silovsky, Stemmer, and
  Vesely]{Povey11_kaldi}
Daniel Povey, Arnab Ghoshal, Gilles Boulianne, Lukas Burget, Ondrej Glembek,
  Nagendra Goel, Mirko Hannemann, Petr Motlicek, Yanmin Qian, Petr Schwarz, Jan
  Silovsky, Georg Stemmer, and Karel Vesely.
\newblock {The Kaldi Speech Recognition Toolkit}.
\newblock In \emph{IEEE 2011 Workshop on Automatic Speech Recognition and
  Understanding}. IEEE Signal Processing Society, December 2011.
\newblock IEEE Catalog No.: CFP11SRW-USB.

\bibitem[Pratap et~al.(2020)Pratap, Xu, Sriram, Synnaeve, and
  Collobert]{Pratap_2020}
Vineel Pratap, Qiantong Xu, Anuroop Sriram, Gabriel Synnaeve, and Ronan
  Collobert.
\newblock {MLS}: A large-scale multilingual dataset for speech research.
\newblock In \emph{Interspeech 2020}. {ISCA}, oct 2020.
\newblock \doi{10.21437/interspeech.2020-2826}.
\newblock URL \url{https://doi.org/10.21437%2Finterspeech.2020-2826}.

\bibitem[Radford et~al.(2019)Radford, Wu, Child, Luan, Amodei, and
  Sutskever]{radford19_language}
Alec Radford, Jeff Wu, Rewon Child, David Luan, Dario Amodei, and Ilya
  Sutskever.
\newblock {Language Models are Unsupervised Multitask Learners}.
\newblock Technical report, OpenAI, 2019.
\newblock URL
  \url{https://d4mucfpksywv.cloudfront.net/better-language-models/language-models.pdf}.

\bibitem[Radford et~al.(2022)Radford, Kim, Xu, Brockman, McLeavey, and
  Sutskever]{radford22_whipser}
Alec Radford, Jong~Wook Kim, Tao Xu, Greg Brockman, Christine McLeavey, and
  Ilya Sutskever.
\newblock {Robust Speech Recognition via Large-Scale Weak Supervision}.
\newblock Technical report, OpenAI, 2022.
\newblock URL \url{https://cdn.openai.com/papers/whisper.pdf}.

\bibitem[Renals et~al.(2007)Renals, Hain, and Bourlard]{renals07_ami}
Steve Renals, Thomas Hain, and Herve Bourlard.
\newblock {Recognition and understanding of meetings the AMI and AMIDA
  projects}.
\newblock In \emph{2007 IEEE Workshop on Automatic Speech Recognition and
  Understanding (ASRU)}, pp.\  238--247, 2007.
\newblock \doi{10.1109/ASRU.2007.4430116}.

\bibitem[Sculley et~al.(2015)Sculley, Holt, Golovin, Davydov, Phillips, Ebner,
  Chaudhary, Young, Crespo, and Dennison]{sculley15-hidden-technical-debt}
D.~Sculley, Gary Holt, Daniel Golovin, Eugene Davydov, Todd Phillips, Dietmar
  Ebner, Vinay Chaudhary, Michael Young, Jean-Fran\c{c}ois Crespo, and Dan
  Dennison.
\newblock {Hidden Technical Debt in Machine Learning Systems}.
\newblock In C.~Cortes, N.~Lawrence, D.~Lee, M.~Sugiyama, and R.~Garnett
  (eds.), \emph{Advances in Neural Information Processing Systems}, volume~28.
  Curran Associates, Inc., 2015.
\newblock URL
  \url{https://proceedings.neurips.cc/paper/2015/file/86df7dcfd896fcaf2674f757a2463eba-Paper.pdf}.

\bibitem[Sennrich et~al.(2016)Sennrich, Haddow, and Birch]{sennrich16_nmt}
Rico Sennrich, Barry Haddow, and Alexandra Birch.
\newblock Neural machine translation of rare words with subword units.
\newblock In \emph{Proceedings of the 54th Annual Meeting of the Association
  for Computational Linguistics (Volume 1: Long Papers)}, pp.\  1715--1725,
  Berlin, Germany, August 2016. Association for Computational Linguistics.
\newblock \doi{10.18653/v1/P16-1162}.
\newblock URL \url{https://aclanthology.org/P16-1162}.

\bibitem[{Shen} et~al.(2019){Shen}, {Nguyen}, {Wu}, {Chen}, {Chen}, {Jia},
  {Kannan}, {Sainath}, {Cao}, {Chiu}, {He}, {Chorowski}, {Hinsu}, {Laurenzo},
  {Qin}, {Firat}, {Macherey}, {Gupta}, {Bapna}, {Zhang}, {Pang}, {Weiss},
  {Prabhavalkar}, {Liang}, {Jacob}, {Liang}, {Lee}, {Chelba}, {Jean}, {Li},
  {Johnson}, {Anil}, {Tibrewal}, {Liu}, {Eriguchi}, {Jaitly}, {Ari}, {Cherry},
  {Haghani}, {Good}, {Cheng}, {Alvarez}, {Caswell}, {Hsu}, {Yang}, {Wang},
  {Gonina}, {Tomanek}, {Vanik}, {Wu}, {Jones}, {Schuster}, {Huang}, {Chen},
  {Irie}, {Foster}, {Richardson}, {Macherey}, {Bruguier}, {Zen}, {Raffel},
  {Kumar}, {Rao}, {Rybach}, {Murray}, {Peddinti}, {Krikun}, {Bacchiani},
  {Jablin}, {Suderman}, {Williams}, {Lee}, {Bhatia}, {Carlson}, {Yavuz},
  {Zhang}, {McGraw}, {Galkin}, {Ge}, {Pundak}, {Whipkey}, {Wang}, {Alon},
  {Lepikhin}, {Tian}, {Sabour}, {Chan}, {Toshniwal}, {Liao}, {Nirschl}, and
  {Rondon}]{shen19_lingvo}
Jonathan {Shen}, Patrick {Nguyen}, Yonghui {Wu}, Zhifeng {Chen}, Mia~X. {Chen},
  Ye~{Jia}, Anjuli {Kannan}, Tara {Sainath}, Yuan {Cao}, Chung-Cheng {Chiu},
  Yanzhang {He}, Jan {Chorowski}, Smit {Hinsu}, Stella {Laurenzo}, James {Qin},
  Orhan {Firat}, Wolfgang {Macherey}, Suyog {Gupta}, Ankur {Bapna}, Shuyuan
  {Zhang}, Ruoming {Pang}, Ron~J. {Weiss}, Rohit {Prabhavalkar}, Qiao {Liang},
  Benoit {Jacob}, Bowen {Liang}, HyoukJoong {Lee}, Ciprian {Chelba},
  S{\'e}bastien {Jean}, Bo~{Li}, Melvin {Johnson}, Rohan {Anil}, Rajat
  {Tibrewal}, Xiaobing {Liu}, Akiko {Eriguchi}, Navdeep {Jaitly}, Naveen {Ari},
  Colin {Cherry}, Parisa {Haghani}, Otavio {Good}, Youlong {Cheng}, Raziel
  {Alvarez}, Isaac {Caswell}, Wei-Ning {Hsu}, Zongheng {Yang}, Kuan-Chieh
  {Wang}, Ekaterina {Gonina}, Katrin {Tomanek}, Ben {Vanik}, Zelin {Wu}, Llion
  {Jones}, Mike {Schuster}, Yanping {Huang}, Dehao {Chen}, Kazuki {Irie},
  George {Foster}, John {Richardson}, Klaus {Macherey}, Antoine {Bruguier},
  Heiga {Zen}, Colin {Raffel}, Shankar {Kumar}, Kanishka {Rao}, David {Rybach},
  Matthew {Murray}, Vijayaditya {Peddinti}, Maxim {Krikun}, Michiel A.~U.
  {Bacchiani}, Thomas~B. {Jablin}, Rob {Suderman}, Ian {Williams}, Benjamin
  {Lee}, Deepti {Bhatia}, Justin {Carlson}, Semih {Yavuz}, Yu~{Zhang}, Ian
  {McGraw}, Max {Galkin}, Qi~{Ge}, Golan {Pundak}, Chad {Whipkey}, Todd {Wang},
  Uri {Alon}, Dmitry {Lepikhin}, Ye~{Tian}, Sara {Sabour}, William {Chan},
  Shubham {Toshniwal}, Baohua {Liao}, Michael {Nirschl}, and Pat {Rondon}.
\newblock {Lingvo: a Modular and Scalable Framework for Sequence-to-Sequence
  Modeling}.
\newblock \emph{arXiv e-prints}, art. arXiv:1902.08295, February 2019.

\bibitem[Synnaeve et~al.(2020)Synnaeve, Xu, Kahn, Likhomanenko, Grave, Pratap,
  Sriram, Liptchinsky, and Collobert]{synnaeve20_endtoend}
Gabriel Synnaeve, Qiantong Xu, Jacob Kahn, Tatiana Likhomanenko, Edouard Grave,
  Vineel Pratap, Anuroop Sriram, Vitaliy Liptchinsky, and Ronan Collobert.
\newblock "end-to-end asr: from supervised to semi-supervised learning with
  modern architectures.
\newblock In \emph{ICML 2020 Workshop on Self-supervision in Audio and Speech},
  2020.
\newblock URL \url{https://openreview.net/forum?id=OSVxDDc360z}.

\bibitem[{T{\"u}ske} et~al.(2021){T{\"u}ske}, {Saon}, and
  {Kingsbury}]{tuske21_limit}
Zolt{\'a}n {T{\"u}ske}, George {Saon}, and Brian {Kingsbury}.
\newblock {On the limit of English conversational speech recognition}.
\newblock \emph{arXiv e-prints}, art. arXiv:2105.00982, May 2021.

\bibitem[Vaswani et~al.(2017)Vaswani, Shazeer, Parmar, Uszkoreit, Jones, Gomez,
  Kaiser, and Polosukhin]{vaswani17_attention}
Ashish Vaswani, Noam Shazeer, Niki Parmar, Jakob Uszkoreit, Llion Jones,
  Aidan~N Gomez, \L~ukasz Kaiser, and Illia Polosukhin.
\newblock {Attention is All you Need}.
\newblock In I.~Guyon, U.~Von Luxburg, S.~Bengio, H.~Wallach, R.~Fergus,
  S.~Vishwanathan, and R.~Garnett (eds.), \emph{Advances in Neural Information
  Processing Systems}, volume~30. Curran Associates, Inc., 2017.
\newblock URL
  \url{https://proceedings.neurips.cc/paper/2017/file/3f5ee243547dee91fbd053c1c4a845aa-Paper.pdf}.

\bibitem[Vincent et~al.(2017)Vincent, Watanabe, Nugraha, Barker, and
  Marxer]{chime4_17}
Emmanuel Vincent, Shinji Watanabe, Aditya~Arie Nugraha, Jon Barker, and Ricard
  Marxer.
\newblock An analysis of environment, microphone and data simulation mismatches
  in robust speech recognition.
\newblock \emph{Comput. Speech Lang.}, 46\penalty0 (C):\penalty0 535–557, nov
  2017.
\newblock ISSN 0885-2308.
\newblock \doi{10.1016/j.csl.2016.11.005}.
\newblock URL \url{https://doi.org/10.1016/j.csl.2016.11.005}.

\bibitem[Wang et~al.(2019{\natexlab{a}})Wang, Pruksachatkun, Nangia, Singh,
  Michael, Hill, Levy, and Bowman]{wang19_superglue}
Alex Wang, Yada Pruksachatkun, Nikita Nangia, Amanpreet Singh, Julian Michael,
  Felix Hill, Omer Levy, and Samuel Bowman.
\newblock {SuperGLUE: A Stickier Benchmark for General-Purpose Language
  Understanding Systems}.
\newblock In H.~Wallach, H.~Larochelle, A.~Beygelzimer, F.~d\textquotesingle
  Alch\'{e}-Buc, E.~Fox, and R.~Garnett (eds.), \emph{Advances in Neural
  Information Processing Systems}, volume~32. Curran Associates, Inc.,
  2019{\natexlab{a}}.
\newblock URL
  \url{https://proceedings.neurips.cc/paper/2019/file/4496bf24afe7fab6f046bf4923da8de6-Paper.pdf}.

\bibitem[Wang et~al.(2019{\natexlab{b}})Wang, Singh, Michael, Hill, Levy, and
  Bowman]{wang19_glue}
Alex Wang, Amanpreet Singh, Julian Michael, Felix Hill, Omer Levy, and
  Samuel~R. Bowman.
\newblock {GLUE: A Multi-Task Benchmark and Analysis Platform for Natural
  Language Understanding}.
\newblock In \emph{7th International Conference on Learning Representations,
  {ICLR} 2019, New Orleans, LA, USA, May 6-9, 2019}. OpenReview.net,
  2019{\natexlab{b}}.
\newblock URL \url{https://openreview.net/forum?id=rJ4km2R5t7}.

\bibitem[Wang et~al.(2021)Wang, Riviere, Lee, Wu, Talnikar, Haziza, Williamson,
  Pino, and Dupoux]{wang21_voxpopuli}
Changhan Wang, Morgane Riviere, Ann Lee, Anne Wu, Chaitanya Talnikar, Daniel
  Haziza, Mary Williamson, Juan Pino, and Emmanuel Dupoux.
\newblock {VoxPopuli: A Large-Scale Multilingual Speech Corpus for
  Representation Learning, Semi-Supervised Learning and Interpretation}.
\newblock In \emph{Proceedings of the 59th Annual Meeting of the Association
  for Computational Linguistics and the 11th International Joint Conference on
  Natural Language Processing (Volume 1: Long Papers)}, pp.\  993--1003,
  Online, August 2021. Association for Computational Linguistics.
\newblock \doi{10.18653/v1/2021.acl-long.80}.
\newblock URL \url{https://aclanthology.org/2021.acl-long.80}.

\bibitem[wen Yang et~al.(2021)wen Yang, Chi, Chuang, Lai, Lakhotia, Lin, Liu,
  Shi, Chang, Lin, Huang, Tseng, tik Lee, Liu, Huang, Dong, Li, Watanabe,
  Mohamed, and yi~Lee]{yang21_superb}
Shu wen Yang, Po-Han Chi, Yung-Sung Chuang, Cheng-I~Jeff Lai, Kushal Lakhotia,
  Yist~Y. Lin, Andy~T. Liu, Jiatong Shi, Xuankai Chang, Guan-Ting Lin,
  Tzu-Hsien Huang, Wei-Cheng Tseng, Ko~tik Lee, Da-Rong Liu, Zili Huang, Shuyan
  Dong, Shang-Wen Li, Shinji Watanabe, Abdelrahman Mohamed, and Hung yi~Lee.
\newblock {SUPERB: Speech Processing Universal PERformance Benchmark}.
\newblock In \emph{Proc. Interspeech 2021}, pp.\  1194--1198, 2021.
\newblock \doi{10.21437/Interspeech.2021-1775}.

\bibitem[Wolf et~al.(2020)Wolf, Debut, Sanh, Chaumond, Delangue, Moi, Cistac,
  Rault, Louf, Funtowicz, Davison, Shleifer, von Platen, Ma, Jernite, Plu, Xu,
  Scao, Gugger, Drame, Lhoest, and Rush]{wolf20-transformers}
Thomas Wolf, Lysandre Debut, Victor Sanh, Julien Chaumond, Clement Delangue,
  Anthony Moi, Pierric Cistac, Tim Rault, Rémi Louf, Morgan Funtowicz, Joe
  Davison, Sam Shleifer, Patrick von Platen, Clara Ma, Yacine Jernite, Julien
  Plu, Canwen Xu, Teven~Le Scao, Sylvain Gugger, Mariama Drame, Quentin Lhoest,
  and Alexander~M. Rush.
\newblock {Transformers: State-of-the-Art Natural Language Processing}.
\newblock In \emph{Proceedings of the 2020 Conference on Empirical Methods in
  Natural Language Processing: System Demonstrations}, pp.\  38--45, Online,
  October 2020. Association for Computational Linguistics.
\newblock URL \url{https://www.aclweb.org/anthology/2020.emnlp-demos.6}.

\bibitem[Yamagishi et~al.(2019)Yamagishi, Veaux, and MacDonald]{vctk_19}
Junichi Yamagishi, Christophe Veaux, and Kirsten MacDonald.
\newblock {CSTR VCTK Corpus: English Multi-speaker Corpus for CSTR Voice
  Cloning Toolkit (version 0.92)}.
\newblock 2019.

\bibitem[Yuan \& Briscoe(2016)Yuan and Briscoe]{yuan-briscoe-2016-grammatical}
Zheng Yuan and Ted Briscoe.
\newblock {Grammatical error correction using neural machine translation}.
\newblock In \emph{Proceedings of the 2016 Conference of the North {A}merican
  Chapter of the Association for Computational Linguistics: Human Language
  Technologies}, pp.\  380--386, San Diego, California, June 2016. Association
  for Computational Linguistics.
\newblock \doi{10.18653/v1/N16-1042}.
\newblock URL \url{https://aclanthology.org/N16-1042}.

\bibitem[{Zhang} et~al.(2020){Zhang}, {Qin}, {Park}, {Han}, {Chiu}, {Pang},
  {Le}, and {Wu}]{zhang20_pushing}
Yu~{Zhang}, James {Qin}, Daniel~S. {Park}, Wei {Han}, Chung-Cheng {Chiu},
  Ruoming {Pang}, Quoc~V. {Le}, and Yonghui {Wu}.
\newblock {Pushing the Limits of Semi-Supervised Learning for Automatic Speech
  Recognition}.
\newblock \emph{arXiv e-prints}, art. arXiv:2010.10504, October 2020.

\bibitem[Zhang et~al.(2022)Zhang, Park, Han, Qin, Gulati, Shor, Jansen, Xu,
  Huang, Wang, Zhou, Li, Ma, Chan, Yu, Wang, Cao, Sim, Ramabhadran, Sainath,
  Beaufays, Chen, V.~Le, Chiu, Pang, and Wu]{zhang22_bigssl}
Yu~Zhang, Daniel~S. Park, Wei Han, James Qin, Anmol Gulati, Joel Shor, Aren
  Jansen, Yuanzhong Xu, Yanping Huang, Shibo Wang, Zongwei Zhou, Bo~Li, Min Ma,
  William Chan, Jiahui Yu, Yongqiang Wang, Liangliang Cao, Khe~Chai Sim,
  Bhuvana Ramabhadran, Tara~N. Sainath, Fran\c{c}oise Beaufays, Zhifeng Chen,
  Quoc V.~Le, Chung-Cheng Chiu, Ruoming Pang, and Yonghui Wu.
\newblock {BigSSL: Exploring the Frontier of Large-Scale Semi-Supervised
  Learning for Automatic Speech Recognition}.
\newblock \emph{IEEE Journal of Selected Topics in Signal Processing}, pp.\
  1--14, 2022.
\newblock \doi{10.1109/JSTSP.2022.3182537}.

\end{thebibliography}
\bibliographystyle{iclr2023_conference}

\appendix
\section{Additional Dataset Details} \label{sec:appendix-data-preparation}

In this section, we give an exhaustive list of the relevant qualitative and quantitative information regarding the datasets used in ESB. Furthermore, we list the minimal transcription error corrections that were performed on the raw transcriptions, in order to ensure systems were trained on suitably formatted text. Finally, we present a diagnostic dataset to help people submitting to ESB find weaknesses in their systems.

\subsection{In-detail datasets information}
In the following section, we give a detailed overview of all relevant statistics for each dataset in ESB.
The data is summarised in Tables~\ref{tab:datasets-summary-full-part-1} and ~\ref{tab:datasets-summary-full-part-2}.
For some datasets, we could not reliably retrieve the number of speakers. We denote these missing entries with \emph{?}.
In addition to the metrics below, we attempted to estimate the signal-to-noise ratio (SNR) of each dataset to quantify their \emph{noisiness}. We experimented with WADA-SNR \citep{wada_19} using an open-source implementation\footnote{SNR: \url{https://gist.github.com/johnmeade/d8d2c67b87cda95cd253f55c21387e75}}, but found our results to be inconclusive with extremely high variance within datasets, and thus deemed them to be largely incorrect.
In addition, we tried pre-processing the audio with voice-activity detection (VAD), using the popular webrtcvad\footnote{webrtcvad: \url{https://github.com/wiseman/py-webrtcvad}} VAD tool. However, this also produced unreasonable estimates, again with very high variance within datasets. Combined with a lack of literature on reference numbers, we excluded SNR estimations from our results.

\begin{table}
\caption{Exhaustive datasets description and statistics (Part 1). Metrics in sample numbers are denoted by (\#). Missing speaker number entries are marked with \emph{?}. Sampling rate is measured in kilohertz (kHz).}
\label{tab:datasets-summary-full-part-1}
\begin{center}
\begin{tabular}{lllll} 
\toprule
                & \textbf{Domain}              & \textbf{Rec. device} & \textbf{Source}     & \textbf{Speakers (\#)}     \\ 
\midrule
LibriSpeech (c) & Audiobook                    & Close-talk mic.          & Expert              & {1252}   \\
LibriSpeech (o) & Audiobook                    & Close-talk mic.          & Expert              & {1232}   \\
Common Voice    & Wikipedia                    & Teleconf.                & Crowd                  & {81085}  \\
VoxPopuli       & EU Parliament                & Close-talk mic.          & Expert              & {1313}   \\
TED-LIUM        & TED talks                    & Close-talk mic.          & Expert              & {2028}   \\
GigaSpeech      & Audiobook, pod., YouT.    & Close-talk mic.          & Expert, Crowd         & {?}      \\
SPGISpeech      & Financial Meet.              & Teleconf.                & Expert              & {50000}  \\
Earnings-22     & Financial Meet.              & Teleconf.                & Expert              & {?}      \\
AMI             & Meetings                     & Close-talk mic.          & Expert              & {?}      \\
Switchboard     & Telephone conv.              & Teleconf.                & Expert              & {543}    \\
CHiME-4         & Broadcast news               & Distant Mic.             & Expert              & {87}     \\ 
\toprule
                & \textbf{Style}               & \textbf{Non-native}      & \textbf{Alignment}  & \textbf{License}           \\ 
\midrule
LibriSpeech (c) & Narrated                     & No                       & Automatic           & CC-BY-4.0                  \\
LibriSpeech (o) & Narrated                     & No                       & Automatic           & CC-BY-4.0                  \\
Common Voice    & Narrated                     & Yes                      & Manual              & CC0-1.0                    \\
VoxPopuli       & Oratory                      & Yes                      & Automatic           & CC0                        \\
TED-LIUM        & Oratory                      & Yes                      & Automatic           & CC-BY-NC-ND 3.0            \\
GigaSpeech      & Narrated, spontaneous        & Yes                      & Automatic           & apache-2.0                 \\
SPGISpeech      & Oratory, spontaneous         & Yes                      & Manual              & User Agreement      \\
Earnings-22     & Oratory, spontaneous         & Yes                      & Automatic           & CC-BY-SA-4.0               \\
AMI             & Spontaneous                  & Yes                      & Automatic           & CC-BY-4.0                  \\
Switchboard     & Spontaneous                  & No                       & Manual              & LDC         \\
CHiME-4         & Narrated                    & No                       & Automatic           & LDC         \\ 
\toprule
                & \textbf{Samp. Rate (kHz)} & \textbf{Cased}           & \textbf{Punctuated} & \textbf{Orthographic}      \\ 
\midrule
LibriSpeech (c) & {16}       & No                       & No                  & No                         \\
LibriSpeech (o) & {16}       & No                       & No                  & No                         \\
Common Voice    & {48}       & Yes                      & Yes                 & No                         \\
VoxPopuli       & {16}       & No                       & No                  & No                         \\
TED-LIUM        & {16}       & No                       & No                  & No                         \\
GigaSpeech      & {16}       & No                       & Yes                 & No                         \\
SPGISpeech      & {16}       & Yes                      & Yes                 & Yes                        \\
Earnings-22     & {16}       & Yes                      & Yes                 & Yes                        \\
AMI             & {16}       & Yes                      & Yes                 & Yes                        \\
Switchboard     & {8}        & No                       & No                  & No                         \\
CHiME-4         & {16}       & Yes                      & Yes                 & Yes                        \\
\bottomrule
\end{tabular}
\end{center}
\end{table}

\begin{table}
\caption{Exhaustive datasets description and statistics (Part 2). Metrics in sample numbers are denoted by (\#). Metrics in hours are denoted by (h). Metrics in seconds are denoted by (s). Metrics in word length are denoted by (word).}
\label{tab:datasets-summary-full-part-2}
\begin{center}
\begin{tabular}{lrrrr} 
\toprule
                & \textbf{Validation (h)} & \textbf{Validation (h)}  & \textbf{Test (h)}  & \textbf{Mean Length (s)}      \\ 
\midrule
LibriSpeech (c) & 460                     & 5                        & 5                  & 12.4                   \\
LibriSpeech (o) & 500                     & 5                        & 5                  & 11.8                   \\
Common Voice    & 1409                    & 27                       & 27                 & 5.6                    \\
VoxPopuli       & 523                     & 5                        & 5                  & 10.3                   \\
TED-LIUM        & 454                     & 2                        & 3                  & 6.1                    \\
GigaSpeech      & 2500                    & 12                       & 40                 & 4.0                    \\
SPGISpeech      & 4900                    & 100                      & 100                & 9.2                    \\
Earnings-22     & 105                     & 5                        & 5                  & 7.2                    \\
AMI             & 78                      & 9                        & 9                  & 2.6                    \\
Switchboard     & 3572                    & 30                       & 7                  & 3.5                    \\
CHiME-4         & 19                      & 11                       & 7                  & 6.7                    \\ 
\toprule
                & \textbf{Train (\#)}     & \textbf{Validation (\#)} & \textbf{Test (\#)} & \textbf{Mean Length (words)}  \\ 
\midrule
LibriSpeech (c) & 132,553                 & 2,703                    & 2,620              & 34.0                   \\
LibriSpeech (o) & 148,688                 & 2,864                    & 2,939              & 31.9                   \\
Common Voice    & 890,116                 & 16,335                   & 16,335             & 9.9                    \\
VoxPopuli       & 182,482                 & 1,753                    & 1,842              & 26.1                   \\
TED-LIUM        & 268,263                 & 591                      & 1,469              & 18.3                   \\
GigaSpeech      & 2,266,371               & 6,750                    & 25,619             & 12.9                   \\
SPGISpeech      & 1,926,805               & 39,304                   & 39,341             & 24.1                   \\
Earnings-22     & 52,006                  & 2,650                    & 2,735              & 17.6                   \\
AMI             & 108,502                 & 13,098                   & 12,643             & 7.3                    \\
Switchboard     & 3,712,270               & 21,296                   & 4,466              & 9.9                    \\
CHiME-4         & 9137                    & 6426                     & 4096               & 16.3                   \\
\bottomrule
\end{tabular}
\end{center}
\end{table}

\subsection{Transcription error correction}\label{sub:appendix-error-correction}
Below we describe the annotation correction steps taken for each dataset. To re-iterate, we do not consider any of the following pre-processing steps to be any form of \emph{normalisation}. Instead, we see them as steps to correct erroneously annotated transcriptions or to remove \emph{junk annotations}, such as $<$unk$>$ or $<$noise$>$. These junk annotations fully unrelated to any kind of punctuation or text and cannot be considered part of speech recognition. All our error corrections steps are reflected in the publicly available code\footnote{ESB Datasets: \url{https://huggingface.co/datasets/esb/esb-datasets}} that is used to download and prepare the benchmark's datasets.

\paragraph{LibriSpeech}
No annotation error corrections.

\paragraph{Common Voice} Many examples have incorrect trailing quotations marks, e.g \emph{"the cat sat on the mat."} instead of \emph{the cat sat on the mat.}, probably due to wrong transcription submissions. It does not make sense to wrap a standalone sentence that is considered without any context into quotation marks. In these cases, we strip the trailing quotation marks, leaving: \emph{the cat sat on the mat.}. Additionally double or triple quotation marks are corrected to single quotation marks (e.g. \emph{"""wait!""" they cried} to \emph{"wait!" they cried}) as double or triple quotation marks do not exist in the English language.

\paragraph{VoxPopuli}
No annotation error corrections.

\paragraph{TED-LIUM}
Transcriptions in the train set contain instances of the \emph{$<$unk$>$} token that are not present in the validation and test sets. We remove these tokens from the train set. Additionally, we correct incorrect leading spaces before apostrophes by collapsing spaced apostrophes into un-spaced apostrophes (e.g. \emph{it 's} to \emph{it's}). We omit transcriptions labelled \emph{ignore\_time\_segment\_in\_scoring} from our evaluation by filtering them out.

\paragraph{GigaSpeech}
We remove official junk tokens (\emph{$<$sil$>$}, \emph{$<$music$>$}, \emph{$<$noise$>$}, \emph{$<$other$>$}) as they cannot be considered audio transcriptions, but rather elements useful for audio classification. We convert the spelled out punctuation to symbolic form (e.g. \emph{$<$comma$>$} to \emph{,}) since the speaker did not pronounce \emph{comma}, but instead the orthographic comma is meant.

\paragraph{Earnings-22}
The Earnings-22 dataset contains audio recordings of financial meetings upwards of 10 minutes in duration. We generate time-stamps for the audio files using the official wav2vec 2.0 CTC + $4$-gram model fine-tuned on LibriSpeech \citep{baevski20_wav2vec2}. We split samples at the time-stamps for punctuation. If the split samples are longer than 20 s, we further split them at the longest silence in the utterance. We then train a wav2vec 2.0 CTC system on audio-transcription pairs. We repeat the process of generating time-stamps to yield more robust audio segments.

To form train-validation-test splits, we partition based on audio files, thus keeping speakers distinct between the splits. Files \texttt{4420696.wav,} \texttt{4448760.wav,} \texttt{4461799.wav,} \texttt{4469836.wav,} \texttt{4473238.wav} and \texttt{4482110.wav} form the validation split. Files \texttt{4432298.wav,} \texttt{4450488.wav,} \texttt{4470290.wav,} \texttt{4479741.wav,} \texttt{4483338.wav} and \texttt{4485244.wav} form the test split. The remainder form the train split.

For transcription error correction, we remove the official junk tokens (\emph{$<$crosstalk$>$}, \emph{$<$affirmative$>$}, \emph{$<$inaudible$>$}, \emph{$<$laugh$>$}).

\paragraph{SPGISpeech}
No annotation error corrections.

\paragraph{AMI}
Audio samples in the AMI meeting corpus vary from between 10 and 60 minutes in duration. We segment the audio samples according the the Kaldi \citep{Povey11_kaldi} recipe for AMI\footnote{https://github.com/kaldi-asr/kaldi/tree/master/egs/ami/s5b}; we split samples longer than 30 words at the time-stamps for punctuation to yield utterance of suitable length for training speech recognition systems.

We remove the junk token \emph{$<$unk$>$}, but otherwise leave the transcriptions un-changed. We fully retain the orthography of the text.

\paragraph{SwitchBoard (optional)}
We partition 5\% of the SwitchBoard corpus to form the validation split:\texttt{sw02001-sw02096} and \texttt{sw04300-sw04387} are partitioned as the validation split, the remainder form the train split.

We remove background noises and non-speech sounds denoted by square brackets, for example \emph{[silence]}. We remove angle braced words that mark speech to someone, such as \emph{$<$a\_aside$>$}, \emph{$<$b\_aside$>$}, \emph{$<$e\_aside$>$}. We remove partially pronounced words, again denoted by square brackets, for example \emph{comm[unity]-} is corrected to \emph{comm-}. We remove annotations for common alternate pronunciations denoted by underscores, for instance \emph{okay\_1} is corrected to \emph{okay}. Words that contain laughter are donated in square brackets, e.g.: \emph{[laughter-because]}. We extract the relevant word only: \emph{because}. We remove the curly braces that denote coinages, changing \emph{\{alrighty\}} to \emph{alrighty}. Filler words such as \emph{uh} and \emph{uhm} are annotated in the train set but not the test set. We thus remove these from the train set.

\paragraph{CHiME-4 (optional)}
We convert out all spelled out punctuation tokens to their symbolic form (e.g. \emph{COMMA} to \emph{,}). We do not remove any tokens from the originally annotated transcriptions.

\subsection{Diagnostic Dataset}
We also provide a new diagnostic dataset consisting of re-annotated portions of ESB using a consistent transcription style, including both normalised and orthographic text formats. As such, it facilitates the reliable evaluation across existing academic datasets and encourages the development of new end-to-end ASR systems.

\section{Additional Baseline Details} \label{sec:appendix-baseline}
In this section, we present the five end-to-end baseline systems in more detail. We include details on network topology (architecture, number of layers, dimensions), model initialisation, training and evaluation.

\paragraph{wav2vec 2.0 CTC} A wav2vec 2.0 \citep{baevski20_wav2vec2} encoder consisting of seven layers of convolutions (512 channels with strides (5,2,2,2,2,2,2) and kernel widths (10,3,3,3,3,2,2)) followed by a Transformer \citep{vaswani17_attention} network with 24 layers, model dimension 1,024, inner dimension 4,096 and 16 attention heads. To predict characters, we follow \citet{baevski20_wav2vec2} in appending a randomly initialised linear layer to the output of the Transformer block with dimensionality equal to the size of the vocabulary. The wav2vec 2.0 model is implemented as a Flax \citep{heek20_flax} neural network module in the Hugging Face Transformers \citep{wolf20-transformers} library.

We initialise the encoder weights with the official wav2vec 2.0 LARGE checkpoint trained on LibriVox (LV-60k) \citep{baevski20_wav2vec2}. We define the output vocabulary by computing the frequency of characters in the train set and discarding those below a relative frequency of 0.01\%.

For training, we filter audio samples longer than 20 s. We resample all audio data to 16 kHz and normalise utterances to zero mean and unit variance. The system is fine-tuned using the Lingvo \citep{shen19_lingvo} JAX implementation of the CTC objective. During fine-tuning, we follow the settings of \citet{baevski20_wav2vec2} and freeze the parameters of the convolutional waveform encoder. We use an Optax \citep{deepmind20_jax} implementation of the Adam \citep{diederik15_adam} optimiser. We train on a single TPU v3-8 \citep{jouppi20_tpu} with a batch size of 8 sequences per device, giving an effective batch size of 64 sequences. We train for a total of 50k optimisation steps. We use the slanted triangular learning rate (STLR) \citep{howard18_universal} schedule, linearly increasing the learning rate from zero to a maximum of 1e-4 over the first 5k steps and then linearly decaying it to zero. During training, we evaluate the system on the validation set at 10k step intervals. We select the checkpoint with the best validation performance for evaluation on the test set.

\paragraph{wav2vec 2.0 CTC + $\boldsymbol{n}$-gram} We combine the wav2vec 2.0 CTC system with a $5$-gram language model. We use the training transcriptions as an LM corpus for each dataset. We compute a MLE of the $5$-gram KenLM parameters with Kneser-Ney smoothing \citep{ney94_kneser_ney, heafield13_scalable}. For decoding, we use an LM weight of 0.5 and a word-insertion penalty of 1.5. We use 100 beams and a one pass beam-search decoder from pyctcdecode\footnote{\url{https://github.com/kensho-technologies/pyctcdecode}} to perform LM boosted beam search decoding for CTC.

\paragraph{wav2vec 2.0 AED} We employ an attention-based encoder-decoder (AED) system. The encoder uses the same wav2vec 2.0 network as described in the wav2vec 2.0 CTC baseline. The decoder is also a Transformer network, consisting of 12 layers, model dimension 1,024, inner dimension 4,096 and 16 attention heads. We follow \citet{li20-lna} and \citet{babu21-xlsr} in adding a randomly initialised adapter network to interface the encoder and decoder, consisting of three 1-dimensional CNN blocks, each of kernel size 3 and stride 2. The wav2vec 2.0 AED model is implemented as a Flax neural network module in the Hugging Face Transformers library.

We initialise the encoder weights with the official wav2vec 2.0 LARGE LV-60k checkpoint and the decoder model weights with the official BART LARGE \citep{lewis20_bart} checkpoint. The vocabulary is un-changed from the vocabulary of the pretrained BART large model, and thus inherits the BART byte-level Byte-Pair-Encoding (BPE) \citep{sennrich16_nmt} tokenizer.

The wav2vec 2.0 AED system is fine-tuned in much the same way as the CTC system, with the exception of the objective function and learning-rate schedule. The system is fine-tuned using the cross-entropy objective implementation in Optax. We again use the STLR schedule, linearly increasing the learning rate from zero to a maximum of 3e-4 over the first 500 steps and then linearly decaying it to zero. We select the checkpoint with the best validation set performance for optimising the generation hyper-parameters. We use a beam size of 12 and a maximum sequence length of 225 tokens. We select the length penalty as the value that yields the best performance on the validation set. We use this setting for the final evaluation on the test set.

\paragraph{Whisper AED} We employ a second AED network. We use 80-dimensional filterbank features from a 25 ms sliding window and a stride of 10 ms as the inputs to the encoder. The input is passed through two convolutional layers with filter widths of 3 and strides of 1 and 2. The encoder consists of a Transformer network with 12 layers, model dimension 1,024, inner dimension 4,096 and 16 attention heads. The decoder is also a Transformer network with the same dimensions and number of layers. The model is implemented as a PyTorch \citep{paszke19_pytorch} neural network in the official Whisper \citep{radford22_whipser} repository.

We initialise the system weights entirely with the official Whisper medium.en checkpoint pretrained on 680k hours of weakly labelled audio data. The tokenizer is the same BPE tokenizer used in GPT-2 \citep{radford19_language}.

For training, we truncate audio samples to 30 seconds and resample them to 16 kHz. The system is fine-tuned using the cross-entropy objective. During fine-tuning, we freeze the encoder network. We use the PyTorch implementation of the Adam optimiser. We train on a single NVIDIA A-100 GPU \citep{choquette21_a100} with a batch-size of 64. We train for a total of 5k steps. We use the STLR schedule, linearly increasing the learning rate from zero to a maximum of 1e-4 over the first 500 steps and then linearly decaying it to zero. During training, we evaluate the system on the validation set at 500 step intervals. We decode using greedy search with a maximum sequence length of 225 tokens. We select the checkpoint with the best validation performance for evaluation on the test set.

Due to the short period of time between the official Whisper checkpoint release and the submission deadline, we did not exhaustively explore training configurations or generation hyper-parameters (such as beam search). Doing so would most likely have led to improved results. We leave this as future work.

\paragraph{Conformer RNN-T} We use 80-dimensional filterbank features from a 25 ms sliding window and a stride of 10 ms as the inputs to the encoder. The encoder consists of a Conformer network with 24 layers, model dimension 1,024, inner dimension 4096, convolutional kernel size 5 and 8 attention heads. The prediction network consists of 2 RNN layers with hidden dimension 640. The transcription network consists of a single feedforward layer with hidden dimension 640. The unit of prediction for the system is SentencePiece \citep{kudo18_sentencepiece} tokenized text with a vocabulary of size 1,024. The model is implemented as a PyTorch neural network module in the NVIDA NeMo \citep{kuchaiev19_nemo} library.

Since the official weights for the Conformer Transducer are not open-sourced, we use the nearest like-for-like open-source replacement. We initialise the model weights from the NVIDIA NeMo XLARGE checkpoint\footnote{\url{https://catalog.ngc.nvidia.com/orgs/nvidia/teams/nemo/models/stt_en_conformer_transducer_xlarge}} trained on combination of 11 speech recognition datasets totalling nearly 24k hours: LibriSpeech \citep{panayotov15_libripseech}, Fisher \citep{ldc04_fisher_speech, ldc04_fisher_transcriptions}, SwitchBoard \citep{godfrey92_switchboard}, Wall-Street Journal 0 and 1 \citep{ldc93_wsj}, National Speech Corpus (Part 1, Part 6) \citep{national_speech_19}, VCTK \citep{vctk_19}, VoxPopuli \citep{wang21_voxpopuli}, Europarl-ASR (EN) \citep{europarlasr2021}, Multilingual Librispeech (MLS EN, 2k hours subset) \citep{Pratap_2020}, Mozilla Common Voice (version 8.0) \citep{ardila20_commonvoice} and People's Speech (12k hours subset) \citep{galvez21_peoples_speech}. We train a BPE SentencePiece tokenizer on the transcriptions from the train split for each dataset.

For training, we filter audio samples longer than 20 s. We resample all audio data to 16 kHz and normalise utterances to zero mean and unit variance. We use SpecAugment \citep{park19_specaug} for data augmentation during training with mask parameter $F=27$ and ten time masks with maximum time mask ration of $p = 0.05$. We set the maximum size of the time mask to $p$ times the length of the utterance and do not use time warping. We fine-tune the system using the NeMo implementation of the Transducer objective. We use the PyTorch implementation of the Adam optimiser. We train on a single NVIDIA A-100 GPU with a batch size of 8 sequences. We train for a total of 100k optimisation steps. We use the STLR schedule, linearly increasing the learning rate from zero to a maximum of 1e-4 over the first 500 steps and then linearly decaying it to zero. During training, we evaluate the system on the validation set at 2.5k step intervals. We decode using greedy search. We find that this yields comparable results to beam-search with a beam-size of 5, but with substantially faster computation times. We select the checkpoint with the best validation performance for evaluation on the test set.
\section{Development Set Results}
To provide a reference for system development and future work on ESB, we present the best validation set results achieved by our baselines in Table~\ref{tab:dev-set-results}.

\begin{table}[t]
\caption{Baseline performance on the validation sets and overall benchmark scores. We report orthographic WERs in \%. SwitchBoard and CHiME-4 are optional datasets for ESB.}
\label{tab:dev-set-results}
\begin{center}
\begin{tabular}{l>{\raggedleft\arraybackslash}p{1.25cm}>{\raggedleft\arraybackslash}p{1.25cm}>{\raggedleft\arraybackslash}p{1.25cm}>{\raggedleft\arraybackslash}p{1.25cm}>{\raggedleft\arraybackslash}p{1.25cm}}
\toprule
& \multicolumn{3}{c}{\textbf{wav2vec 2.0}} & \multicolumn{1}{c}{\textbf{Whisper}} &  \multicolumn{1}{c}{\textbf{Conformer}} \\
Dataset       & \multicolumn{1}{c}{\textbf{CTC}} & \multicolumn{1}{c}{$\substack{\textbf{CTC +}\\ \text{n-gram}}$} & \multicolumn{1}{c}{\textbf{AED}} & \multicolumn{1}{c}{\textbf{AED}} & \multicolumn{1}{c}{\textbf{RNN-T}}  \\ 
\midrule
LibriSpeech \textit{test-clean} & 2.9                              & 2.4                                                             & 2.8                              & 2.2                                  & 2.0                                     \\
LibriSpeech \textit{test-other} & 7.5                              & 5.9                                                             & 5.8                              & 5.2                                  & 4.0                                     \\
Common Voice                    & 22.8                             & 18.9                                                            & 14.0                             & 13.6                                 & 13.2                                    \\
VoxPopuli                       & 11.4                             & 9.1                                                             & 10.2                             & 7.2                                  & 7.5                                     \\
TED-LIUM                        & 8.7                              & 7.1                                                             & 12.4                             & 5.0                                  & 5.6                                     \\
GigaSpeech                      & 25.9                             & 22.5                                                            & 22.2                             & 18.0                                 & 19.0                                    \\
SPGISpeech                      & 8.1                              & 7.1                                                             & 5.4                              & 5.6                                  & 6.3                                     \\
Earnings-22                     & 26.7                             & 38.9                                                            & 22.8                             & 16.0                                 & 18.3                                    \\
AMI                             & 32.1                             & 32.9                                                            & 20.5                             & 16.5                                 & 17.8                                    \\ 
\midrule
SwitchBoard                     & 15.0                             & 11.4                                                            & 11.3                             & 8.2                                  & 8.3                                     \\
CHiME-4                         & 19.6                             & 18.9                                                            & 52.3                             & 9.1                                  & 12.2                                    \\
\bottomrule
\end{tabular}
\end{center}
\end{table}
\section{Additional Analysis Results} \label{sec:appendix-score-ablation}
Table~\ref{tab:all-baseline-results-ablated} details the orthographic WER scores for the ESB test sets on a per-dataset level for no post-processing, and under three post-processing conditions: (i) remove punctuation, (ii) remove casing, (iii) apply full normalisation. Table~\ref{tab:all-optional-baseline-results-ablated} shows the same metrics for the optional test sets.

\begin{table}[t]
\caption{The effect of punctuation, casing and full normalisation on orthographic WER scores for the ESB test sets. We show the WER without post-processing, WER with punctuation removed, WER with casing removed and WER with full normalisation.}
\label{tab:all-baseline-results-ablated}
\begin{center}
\begin{tabular}{l>{\raggedleft\arraybackslash}p{1.25cm}>{\raggedleft\arraybackslash}p{1.25cm}>{\raggedleft\arraybackslash}p{1.25cm}>{\raggedleft\arraybackslash}p{1.25cm}>{\raggedleft\arraybackslash}p{1.25cm}}
\toprule
& \multicolumn{3}{c}{\textbf{wav2vec 2.0}} & \multicolumn{1}{c}{\textbf{Whisper}} &  \multicolumn{1}{c}{\textbf{Conformer}} \\
Dataset       & \multicolumn{1}{c}{\textbf{CTC}} & \multicolumn{1}{c}{$\substack{\textbf{CTC +}\\ \text{n-gram}}$} & \multicolumn{1}{c}{\textbf{AED}} & \multicolumn{1}{c}{\textbf{AED}} & \multicolumn{1}{c}{\textbf{RNN-T}}  \\ 
\midrule
LibriSpeech \textit{test-clean} & 2.9                              & 2.4                                                             & 2.8                              & 2.2                                  & 2.0                                     \\
- punctuation          & 2.9                              & 2.4                                                             & 2.8                              & 2.2                                  & 2.0                                     \\
- casing               & 2.9                              & 2.4                                                             & 2.8                              & 2.2                                  & 2.0                                     \\
- normalisation        & 2.8                              & 2.3                                                             & 2.6                              & 2.1                                  & 1.9                                     \\ 
\midrule
LibriSpeech \textit{test-other} & 7.5                              & 5.9                                                             & 5.8                              & 5.2                                  & 4.0                                     \\
- punctuation          & 7.5                              & 5.9                                                             & 5.8                              & 5.2                                  & 4.0                                     \\
- casing               & 7.5                              & 5.9                                                             & 5.8                              & 5.2                                  & 4.0                                     \\
- normalisation        & 7.4                              & 5.8                                                             & 5.6                              & 5.1                                  & 3.8                                     \\ 
\midrule
Common Voice           & 26.1                             & 22.2                                                            & 16.3                             & 15.8                                 & 14.8                                    \\
- punctuation          & 23.9                             & 18.8                                                            & 14.4                             & 14.2                                 & 12.3                                    \\
- casing               & 22.4                             & 17.3                                                            & 13.1                             & 12.8                                 & 10.9                                    \\
- normalisation        & 21.9                             & 16.8                                                            & 12.6                             & 12.2                                 & 9.7                                     \\ 
\midrule
VoxPopuli              & 11.4                             & 10.2                                                            & 10.1                             & 7.4                                  & 7.3                                     \\
- punctuation          & 10.3                             & 8.3                                                             & 9.8                              & 7.2                                  & 7.1                                     \\
- casing               & 10.3                             & 8.3                                                             & 9.8                              & 7.2                                  & 7.1                                     \\
- normalisation        & 10.0                             & 8.1                                                             & 9.6                              & 7.0                                  & 6.7                                     \\ 
\midrule
TED-LIUM               & 8.4                              & 6.7                                                             & 6.9                              & 4.7                                  & 5.0                                     \\
- punctuation          & 8.4                              & 6.7                                                             & 6.9                              & 4.7                                  & 5.0                                     \\
- casing               & 8.4                              & 6.7                                                             & 6.9                              & 4.7                                  & 5.0                                     \\
- normalisation        & 7.9                              & 6.2                                                             & 6.3                              & 4.0                                  & 4.5                                     \\ 
\midrule
GigaSpeech             & 25.3                             & 22.0                                                            & 23.4                             & 17.3                                 & 18.6                                    \\
- punctuation          & 18.2                             & 14.9                                                            & 15.3                             & 11.0                                 & 12.4                                    \\
- casing               & 18.2                             & 14.9                                                            & 15.3                             & 11.0                                 & 12.4                                    \\
- normalisation        & 17.5                             & 13.9                                                            & 14.3                             & 10.2                                 & 11.3                                    \\ 
\midrule
SPGISpeech             & 8.1                              & 7.1                                                             & 5.4                              & 5.5                                  & 6.3                                     \\
- punctuation          & 5.8                              & 4.6                                                             & 3.3                              & 3.6                                  & 3.9                                     \\
- casing               & 4.4                              & 3.3                                                             & 2.2                              & 2.4                                  & 2.7                                     \\
- normalisation        & 4.2                              & 3.0                                                             & 2.0                              & 2.2                                  & 2.4                                     \\ 
\midrule
Earnings-22            & 26.0                             & 31.7                                                            & 23.6                             & 16.0                                 & 17.6                                    \\
- punctuation          & 20.9                             & 21.0                                                            & 20.1                             & 12.1                                 & 13.3                                    \\
- casing               & 20.4                             & 20.5                                                            & 19.6                             & 11.5                                 & 12.6                                    \\
- normalisation        & 19.3                             & 18.8                                                            & 18.3                             & 9.9                                  & 10.2                                    \\ 
\midrule
AMI                    & 32.0                             & 33.1                                                            & 19.3                             & 14.5                                 & 15.1                                    \\
- punctuation          & 26.0                             & 25.1                                                            & 16.8                             & 12.2                                 & 12.3                                    \\
- casing               & 24.8                             & 23.7                                                            & 15.5                             & 10.8                                 & 11.0                                    \\
- normalisation        & 23.7                             & 22.0                                                            & 15.0                             & 10.3                                 & 10.2                                    \\ 
\bottomrule
\end{tabular}
\end{center}
\end{table}

\begin{table}[t]
\caption{The effect of punctuation, casing and full normalisation on orthographic WER scores for the optional test sets. We show the WER without post-processing, WER with punctuation removed, WER with casing removed and WER with full normalisation.}
\label{tab:all-optional-baseline-results-ablated}
\begin{center}
\begin{tabular}{l>{\raggedleft\arraybackslash}p{1.25cm}>{\raggedleft\arraybackslash}p{1.25cm}>{\raggedleft\arraybackslash}p{1.25cm}>{\raggedleft\arraybackslash}p{1.25cm}>{\raggedleft\arraybackslash}p{1.25cm}}
\toprule
& \multicolumn{3}{c}{\textbf{wav2vec 2.0}} & \multicolumn{1}{c}{\textbf{Whisper}} &  \multicolumn{1}{c}{\textbf{Conformer}} \\
Dataset       & \multicolumn{1}{c}{\textbf{CTC}} & \multicolumn{1}{c}{$\substack{\textbf{CTC +}\\ \text{n-gram}}$} & \multicolumn{1}{c}{\textbf{AED}} & \multicolumn{1}{c}{\textbf{AED}} & \multicolumn{1}{c}{\textbf{RNN-T}}  \\ 
\midrule
SwitchBoard            & 16.1                             & 12.8                                                            & 15.3                             & 10.0                                 & 10.8                                    \\
- punctuation          & 14.1                             & 10.6                                                            & 12.0                             & 8.1                                  & 8.8                                     \\
- casing               & 14.1                             & 10.6                                                            & 12.0                             & 8.1                                  & 8.8                                     \\
- normalisation        & 14.0                             & 10.4                                                            & 10.2                             & 7.8                                  & 8.3                                     \\ 
\midrule
CallHome               & 26.6                             & 20.9                                                            & 24.3                             & 15.9                                 & 23.3                                    \\
- punctuation          & 25.9                             & 20.2                                                            & 19.8                             & 14.4                                 & 22.4                                    \\
- casing               & 25.9                             & 20.2                                                            & 19.8                             & 14.4                                 & 22.4                                    \\
- normalisation        & 25.7                             & 19.9                                                            & 17.1                             & 13.5                                 & 22.1                                    \\ 
\midrule
CHiME-4                 & 29.2                             & 26.8                                                            & 56.9                             & 12.7                                 & 14.2                                    \\
- punctuation          & 28.3                             & 24.4                                                            & 59.2                             & 11.7                                 & 13.3                                    \\
- casing               & 27.4                             & 23.4                                                            & 59.0                             & 11.0                                 & 12.6                                    \\
- normalisation        & 30.7                             & 25.7                                                            & 62.8                             & 11.9                                 & 13.4                                    \\
\bottomrule
\end{tabular}
\end{center}
\end{table}

\end{document}